\pdfoutput=1

\documentclass[11pt]{article}

\usepackage[preprint]{acl}

\usepackage{times}
\usepackage{latexsym}

\usepackage[T1]{fontenc}

\usepackage[utf8]{inputenc}

\usepackage{microtype}

\usepackage{inconsolata}

\usepackage{graphicx}

\usepackage{threeparttable}
\usepackage{booktabs}
\usepackage{multirow}
\usepackage{multicol}
\usepackage{amsmath}
\usepackage{tabularx}
\usepackage{colortbl}

\title{Beyond Exact Match: Semantically Reassessing Event Extraction \\ by Large Language Models}

\author{
 \textbf{Yi-Fan Lu}$^1$,
 \textbf{Xian-Ling Mao}$^1$,
 \textbf{Tian Lan}$^1$,
 \textbf{Heyan Huang}$^1$,
 \textbf{Chen Xu}$^1$,
 \textbf{Xiaoyan Gao}$^2$
\\
 $^1$Beijing Institute of Technology,
 $^2$Beijing University of Technology
\\
\texttt{\href{mailto:yifanlu@bit.edu.cn}{yifanlu@bit.edu.cn}},
\texttt{\href{mailto:maoxl@bit.edu.cn}{maoxl@bit.edu.cn}},
\\
\texttt{\href{mailto:lantiangmftby@gmail.com}{lantiangmftby@gmail.com}},
\texttt{\href{mailto:chenxu05037@gmail.com}{chenxu05037@gmail.com}},
\\
\texttt{\href{mailto:hhy63@bit.edu.cn}{hhy63@bit.edu.cn}},
\texttt{\href{gaoxiaoyan@bjut.edu.cn}{gaoxiaoyan@bjut.edu.cn}}
}

\begin{document}
\maketitle
\begin{abstract}
Event extraction has gained extensive research attention due to its broad range of applications. 
However, the current mainstream evaluation method for event extraction relies on token-level exact match, which misjudges numerous semantic-level correct cases. This reliance leads to a significant discrepancy between the evaluated performance of models under exact match criteria and their real performance. 
To address this problem, we propose a reliable and semantic evaluation framework for event extraction, named RAEE, which accurately assesses extraction results at semantic-level instead of token-level. 
Specifically, RAEE leverages large language models (LLMs) as evaluation agents, incorporating an adaptive mechanism to achieve adaptive evaluations for precision and recall of triggers and arguments.
Extensive experiments demonstrate that: 
(1) RAEE achieves a very strong correlation with human judgments; 
(2) after reassessing 14 models, including advanced LLMs, on 10 datasets, there is a significant performance gap between exact match and RAEE. The exact match evaluation significantly underestimates the performance of existing event extraction models, and in particular underestimates the capabilities of LLMs; 
(3) fine-grained analysis under RAEE evaluation reveals insightful phenomena worth further exploration.
The evaluation toolkit of our proposed RAEE is publicly released\footnote{\url{https://github.com/Lyfralston/RAEE}}.
\end{abstract}

\section{Introduction}

\begin{figure}[ht]
    \centering
    \includegraphics[width=\linewidth]{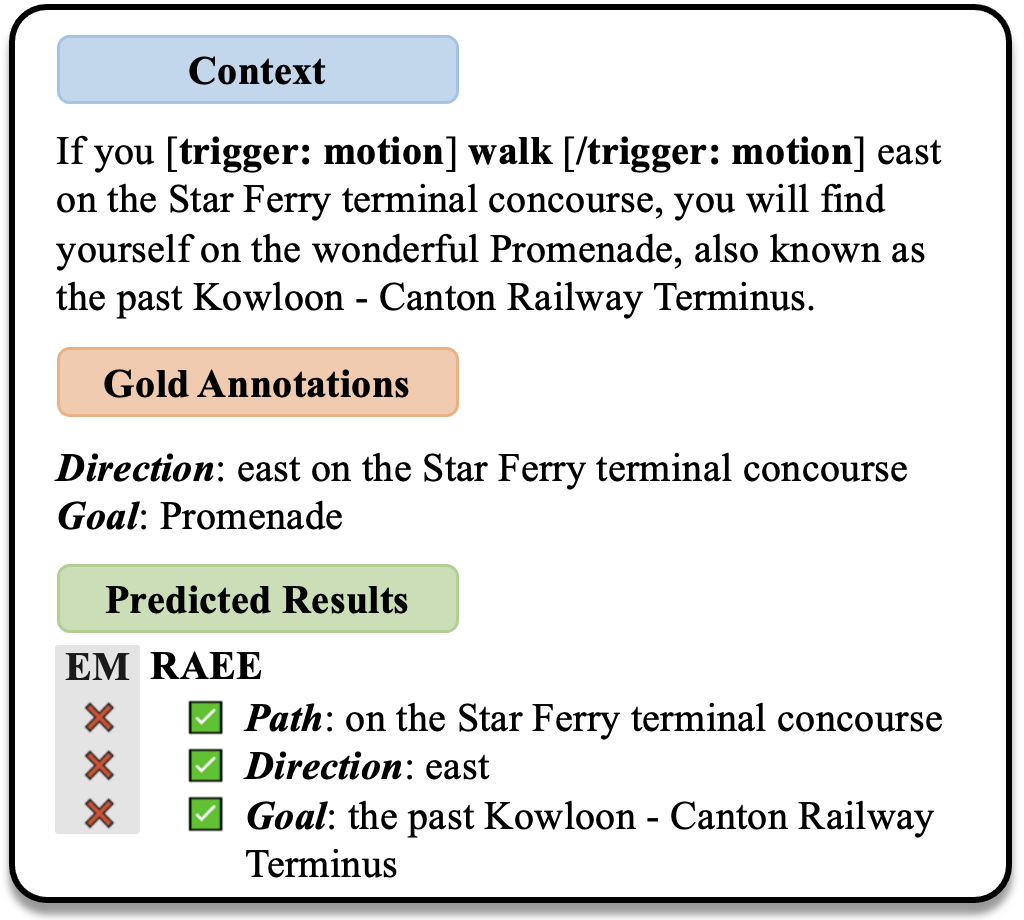}
    \caption{One case of EAE is misjudged by exact match (EM) evaluation but reassessed as correct by our proposed RAEE evaluation. With a given trigger (\textbf{\textit{walk}}), EAE aims to extract its  arguments with semantic roles.}
    \label{fig:intro_case_illustration}
\end{figure}

Event extraction (EE) aims to detect event instances as well as all of their participants and attributes by analyzing and identifying mentions of semantically defined entities and relationships within them \cite{7243219, surveyofee}. It is typically divided into two subtasks, \textbf{E}vent \textbf{D}etection (ED), which focuses on identifying triggers that indicate event occurrence, and \textbf{E}vent \textbf{A}rgument \textbf{E}xtraction (EAE), which extracts arguments with semantic roles related to the trigger.
Since EE benefits a broad range of applications \cite{10.1145/3366423.3380107, han-etal-2021-ester}, it has gained extensive research attention \cite{lu-etal-2022-unified, yan-etal-2023-utc}. 
With the increasing use of large language models (LLMs) as agents \cite{chen2024tevalevaluatingtoolutilization,liu2023agentbenchevaluatingllmsagents}, the requirement for accurate event extraction becomes more important, which helps in effectively tracking and labeling the actions of agents in real-world scenarios \cite{park2023generativeagentsinteractivesimulacra}.


Currently, the most widely-used evaluation method for EE is token-level exact match (EM)\footnote{For the sake of brevity, we will refer to exact match evaluation as EM in the following content.}, which assesses whether the predicted triggers and arguments are identical to the annotated ground-truth in benchmarks~\cite{hsu-etal-2022-degree, lu-etal-2022-unified, hsu-etal-2023-tagprime}. 
However, EM often misjudges a significant number of cases that are semantically correct or convey the same meaning as ground-truth but cannot match it literally~\cite{li-etal-2021-document}.
For example, one case of EAE task, which extracts event arguments corresponding to a given trigger, is illustrated in Figure \ref{fig:intro_case_illustration}. The predicted arguments, \textit{Path: on the Star Ferry terminal concourse} and \textit{Direction: east}, are not the same as ground-truth and are therefore evaluated as failed extractions under EM. However, considering the context, both predictions can be regarded as semantically correct extractions. Also, co-reference instances are misjudged by EM, as shown in the example where \textit{Promenade} and \textit{the past Kowloon - Canton Railway Terminus} refer to the same place.
Consequently, models that may appear to perform poorly under EM actually have strong performance, indicating that the capabilities of existing EE models might be greatly underestimated.
This issue is particularly prominent in generative models, especially LLMs~\cite{huang2024texteebenchmarkreevaluationreflections}, where the generations often cannot exactly match ground-truth.

To address this severe issue, we propose a reliable and semantic evaluation framework designed to \textbf{R}e\textbf{A}ssess \textbf{E}vent \textbf{E}xtraction models at semantic-level instead of token-level, named \textbf{RAEE}.
Specifically, RAEE utilizes advanced LLMs as evaluation agents to achieve automatic assessment. 
By considering the semantics of extractions in the context, the LLM performs binary judgments on the extractions from the perspectives of precision and recall, determining whether they are corrected and whether they are successfully recalled, respectively.
To ensure the reliability of the evaluation, we introduce an adaptive prompting mechanism to enhance LLMs' evaluation consistency with human evaluation.

Experiments begin with a meta-evaluation designed to validate evaluation reliability of RAEE. The results demonstrate that (1) RAEE achieves a strong correlation with human judgments, (2) RAEE significantly outperforms EM and other soft matching evaluation methods, and (3) the proposed adaptive mechanism enhances the reliability of RAEE. 
Then, we apply RAEE to conduct a comprehensive reassessment experiment for 14 EE models on 10 widely-used datasets. 
Extensive experiments reveal that (1) evaluated models exhibit a significant performance gap under EM and RAEE, indicating that existing EE models, particularly generative models, are greatly underestimated, and 
(2) the ranking order among these models shifts after reassessing under RAEE on most datasets, further highlighting the importance of RAEE.
To further explore the reasons behind misjudgments evaluated by EM and failure modes evaluated by RAEE, a fine-grained analysis experiment uncovers insightful phenomena worth future exploration. For example, while most spans are accurately matched, their types are still incorrect under RAEE evaluation.
Finally, we reassess LLMs' event extraction ability and find that LLMs are extremely underestimated by EM, further emphasizing the urgency of incorporating semantic evaluation in the era of LLMs.
In summary, our contributions are as follows:
\begin{itemize}
    \item \textbf{RAEE}: We propose a reliable and semantic evaluation framework to assess event extraction at semantic-level by leveraging advanced LLMs as evaluation agents with an adaptive prompting mechanism. RAEE shows strong correlations with human judgments.
    \item \textbf{Comprehensive evaluation}: We apply our proposed RAEE to comprehensively reassess the performance of previous works and LLMs for event extraction. Experimental results reveal that the capabilities of event extraction models are significantly underestimated, especially of generative models and LLMs.
    \item \textbf{Fine-grained analysis}: We conduct a fine-grained analysis to uncover the 
    reasons behind misjudgments evaluated by EM and failure modes evaluated by RAEE, highlighting areas worth further exploration.
\end{itemize}

\begin{figure*}[h]
    \centering
    \includegraphics[width=\linewidth]{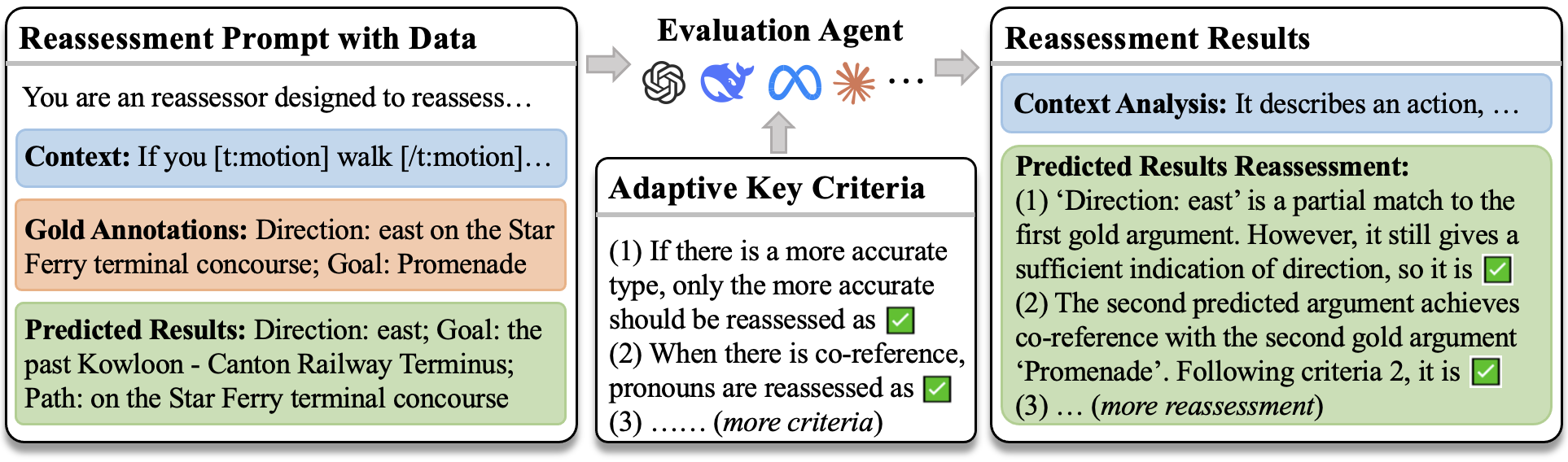}
    \caption{Evaluation process of RAEE, using the precision of EAE as an example (consistent with ED). }
    \label{fig:raee_framework}
\end{figure*}

\section{Related Works}

\subsection{Evaluations on Event Extraction}

Current evaluations of EE typically use precision, recall, and f1-score metrics to evaluate the span and the type of extracted instances, and this approach is consistent across ED and EAE. The majority of research relies on EM evaluation \cite{lin-etal-2020-joint, ma-etal-2022-prompt, hsu-etal-2023-ampere}, and some studies employ language toolkits to apply head noun evaluation \cite{li-etal-2021-document}. Additionally, some works modify the existing evaluation frameworks by incorporating different averaging methods for calculating f1-score \cite{zheng-etal-2021-revisiting} or introducing additional metrics from downstream applications \cite{peng-etal-2023-omnievent}. The consistency and fairness of variable alignment with model outputs are also emphasized in recent studies \cite{peng-etal-2023-devil, huang2024texteebenchmarkreevaluationreflections}.
However, these works uniformly adopt EM when determining span matches, and overlook instances where the semantically correct span cannot exactly match ground-truth. To address this oversight, we propose a semantic-level evaluation framework and provide a more reliable evaluation of model performances.

\subsection{LLMs as Judge Models}

There is an emerging trend in the research community to leverage LLMs for automatically evaluating the quality of generations, thereby reducing the significant cost of human annotations \cite{lan2024criticbench,lin2024criticbench,luo2023critique,fu2023gptscore,wang2023chatgpt}.
Extensive studies \cite{fu2023gptscore,wang2023chatgpt} have shown that advanced LLMs, such as GPT-4, achieve very close correlations with human judgments across various generative NLP tasks, including dialogue generation \cite{fu2023gptscore}, text summarization \cite{liu2023g} and structured generation \cite{wang2023chatgpt}.

In this paper, we utilize advanced LLMs to emulate human annotators in automatically evaluating EE. 
Advanced LLMs closely align with human judgments at a fraction of the cost \cite{lan2020pone}, making our RAEE a scalable and reliable automatic evaluation method for EE.

\section{Semantic Reassessing Framework}
To address the misjudgments caused by EM, we propose a semantic evaluation framework, named RAEE. 
In this section, we will introduce: 
(1) how semantic-level F1-score is calculated in RAEE in Section~\ref{sec:semantic_f1};
(2) the adaptive mechanism of RAEE in Section~\ref{sec:adaptive}; 
(3) the evaluation reliability of RAEE in Section \ref{sec:human_evaluation}.

\subsection{Semantic-level F1-score}
\label{sec:semantic_f1}
Assume there are a piece of context $\bf C$, predicted results from an EE model ${\bf P} = {\rm Model}(\bf C)$, and gold annotations from human ${\bf G} = {\rm Human}(\bf C)$. 
RAEE prompts an evaluation agent LLM to consider the context during evaluation, assessing predictions and annotations separately when calculating precision and recall.
Specifically, RAEE requires the LLM to identify a correct prediction set $\bf P_c$ and a recalled annotation set $\bf G_r$:
$$
{\bf P_c} = {\rm LLM}({\bf P| C, G}), \quad {\bf G_r} = {\rm LLM}({\bf G| C, P}).
$$
Then, semantic-level f1-score can be computed as follows: 
$$
\bf p = \frac{|P_c|}{|P|}, \bf \quad r = \frac{|G_r|}{|G|}, \quad \bf f1_{\rm RAEE} = \frac{2*p*r}{p+r}.
$$

Figure~\ref{fig:raee_framework} presents an example to evaluate predictions ${\bf P}$ in EAE. The LLM is required to first analyze the context and then evaluate each prediction with a reason to get correct predictions $\bf P_c$.

\begin{table*}[h]
    \centering
    \small
    \begin{threeparttable}
    \begin{tabular}{l|cc|cc|cc|cc}
    \toprule
       \multirow{2}{*}{\textbf{Eval. Methods}}  & \multicolumn{2}{c|}{\textbf{ED precision}} & \multicolumn{2}{c|}{\textbf{ED recall}}  & \multicolumn{2}{c|}{\textbf{EAE precision}} & \multicolumn{2}{c}{\textbf{EAE recall}} \\
       ~ & Per. & Sp. & Per. & Sp. & Per. & Sp. & Per. & Sp. \\
    \midrule
       Human Average  & 89.08 & 74.19 & 96.87 & 92.65 & 93.61 & 85.48 & 96.96 & 94.00 \\
    \midrule
    
       \textit{Deepseek-r1}  & \cellcolor{red!10}77.45$_{1.65}$ & \cellcolor{red!30}53.19$_{2.57}$ & \cellcolor{red!30}86.85$_{0.22}$ & \cellcolor{red!30}72.31$_{0.54}$ & \cellcolor{red!10}79.72$_{0.42}$ & \cellcolor{red!10}55.17$_{0.78}$ & \cellcolor{red!30}93.49$_{0.53}$ & \cellcolor{red!30}87.02$_{1.08}$ \\
       ~~~~- w/o. Adaptive  & 72.60$_{0.93}$ & 48.21$_{1.16}$ & \cellcolor{red!10}85.24$_{0.20}$ & 71.13$_{0.33}$ & 76.98$_{0.51}$ & 54.24$_{1.00}$ & \cellcolor{red!10}91.75$_{0.41}$ & \cellcolor{red!10}83.47$_{0.82}$ \\
       \textit{GPT-4o-240513}  & \cellcolor{red!30}78.29$_{0.50}$ & \cellcolor{red!10}50.49$_{2.31}$ & \cellcolor{red!30}86.85$_{0.80}$ & \cellcolor{red!10}71.70$_{1.74}$ & \cellcolor{red!30}84.63$_{0.69}$ & \cellcolor{red!30}64.93$_{2.27}$ & 83.71$_{0.55}$ & 68.81$_{0.54}$ \\
       ~~~~- w/o. Adaptive  & 75.30$_{1.41}$ & 38.33$_{3.80}$ & 83.72$_{1.23}$ & 69.33$_{1.92}$ & 80.56$_{1.32}$ & 57.73$_{2.63}$ & 82.56$_{2.61}$ & 66.88$_{5.00}$ \\
       \textit{Gemini-1.5-pro} & 72.83$_{0.89}$ & 37.93$_{2.26}$ & 51.49$_{0.77}$ & 34.60$_{0.73}$ & 81.35$_{0.61}$ & 55.03$_{1.48}$ & 77.16$_{0.22}$ & 61.81$_{0.44}$ \\
       ~~~~- w/o. Adaptive  & 71.94$_{1.06}$ & 28.14$_{3.12}$ & 48.51$_{1.77}$ & 31.19$_{2.21}$ & 79.57$_{0.89}$ & 49.62$_{2.59}$ & 68.55$_{1.42}$ & 49.74$_{2.37}$ \\
       \textit{Deepseek-v3}  & 66.39$_{1.59}$ & 28.61$_{3.38}$ & 71.47$_{2.51}$ & 54.03$_{3.13}$ & 77.59$_{2.39}$ & 52.78$_{4.41}$ & 76.33$_{0.54}$ & 57.06$_{1.55}$ \\
       ~~~~- w/o. Adaptive  & 66.71$_{1.49}$ & 31.07$_{1.91}$ & 61.87$_{2.95}$ & 44.04$_{3.37}$ & 72.83$_{0.28}$ & 48.77$_{1.71}$ & 72.10$_{1.93}$ & 49.87$_{2.52}$ \\
       \textit{Claude-3.5-sonnet}  & 63.63$_{1.09}$ & 33.98$_{2.96}$ & 71.62$_{0.92}$ & 51.77$_{1.30}$ & 75.61$_{0.57}$ & 48.65$_{1.31}$ & 71.88$_{1.01}$ & 50.75$_{1.29}$ \\
       ~~~~- w/o. Adaptive  & 61.48$_{0.52}$ & 34.83$_{1.47}$ & 64.32$_{3.04}$ & 43.49$_{4.33}$ & 68.91$_{0.19}$ & 44.41$_{1.03}$ & 68.62$_{1.43}$ & 43.95$_{3.09}$ \\
       Similarity  & 50.70 & 12.62 & 84.82 & 71.40 & 63.81 & 16.76 & 75.79 & 54.44 \\
       Head Noun  & 31.93 & 11.80 & 75.74 & 29.97 & 40.98 & 20.07 & 64.39 & 35.98 \\
       Exact Match  & 28.57 & —— & 72.46 & —— & 31.28 & —— & 54.51 & —— \\
    \bottomrule
    \end{tabular}
    \end{threeparttable}
    \caption{Percent agreement (\%) and Spearman correlation ($\times$100) of different evaluation methods with human judgments on four event extraction subtasks. All LLM experiments are repeated three times. All Spearman correlation scores are associated with the $p$-value $<$ 0.05. For each metric in each subtask, we highlight the highest value using a dark shade, and the second-highest value using a light shade.}
    \label{tab:iaa}
\end{table*}


\subsection{Adaptive Mechanism}
\label{sec:adaptive}
Human annotators may have slightly different subjective perspectives when performing semantic evaluation, which could lead to inconsistencies in assessment \cite{lan2024criticbench, kim2023prometheus}. 
Besides, various datasets often have customized annotation rules, resulting in differing evaluation criteria. 
For example, some datasets provide separate annotations for co-references to assist in the annotation of event extraction \cite{li-etal-2021-document} but some do not \cite{ebner-etal-2020-multi, Satyapanich_Ferraro_Finin_2020}. 
To mitigate inconsistent evaluation and achieve adjustable evaluation in RAEE, we manually identify several key criteria of evaluation bias, and prompt the LLM to follow them to assess extractions step by step, as illustrated in Figure~\ref{fig:raee_framework}. 
When an evaluated content meets a specific evaluation criteria, the evaluation agent will explicitly output the corresponding criteria it follows.

These key criteria are adaptive, which allows researchers to tailor the evaluation criteria of RAEE to their specific needs in different scenarios. In the following evaluation experiments, these key criteria remain consistent to achieve consistent evaluation. The default setting of existing criteria can be found in Appendix~\ref{sec:key_criteria}.

\subsection{Reliability of RAEE}
\label{sec:human_evaluation}

To validate the reliability of RAEE and the effectiveness of its adaptive mechanism, we conduct a meta-evaluation to analyze the correlation between RAEE and human judgments.
Specifically, we sample 800 pieces of data in ED and EAE tasks from mainstream event datasets (presented in the main experiment) across various domains, excluding extractions easily verified as correct or recalled by EM.
We apply different LLMs as the evaluation agent to investigate the impact of agent selection on RAEE.
The full prompts are provided in Appendix~\ref{sec:reassess_prompts}.
For human judgments, 3 professional event extraction annotators evaluate the same data using the same prompts as the LLMs.
We also apply EM, head noun evaluation \cite{li-etal-2021-document} and text similarity evaluation with a well-selected threshold $t = 0.5$ \cite{bge_m3} for comparison. 
Two widely-used inter-annotator agreement (IAA) metrics are selected: Percent agreement \cite{zheng2023judgingllmasajudgemtbenchchatbot, Liu2021TowardsFI} and Spearman correlation \cite{lan2024criticbench,fu2023gptscore}. 
The results are presented in Table~\ref{tab:iaa}.

\subsubsection{Overall Correlation}
When using deepseek-r1 as the evaluation agent of RAEE, although there is a noticeable gap between RAEE and human average evaluation (average 0.197 Spearman correlation and 9.75\%  Percent agreement), RAEE still shows strong consistency with human judgments\footnote{These metrics reflect a strong correlation, as validated in past works \cite{lan2020pone,fu2023gptscore}.}, achieving over 77\% Percent agreement and 0.531 Spearman correlation.
Compared to traditional evaluation methods, RAEE significantly outperforms exact match and head noun evaluations, improving Spearman correlation by over 0.351 across all tasks.
Notably, the similarity-based evaluation shows little difference from RAEE on the ED recall task. 
This is because golden triggers to be evaluated in ED recall task are correct and typically single-word mentions. When the evaluation target shifts to model predictions that might be incorrect, as in the precision task, or involves multi-word mentions, as in the EAE task, the performance of similarity-based evaluation significantly declines.

Furthermore, for all evaluation methods, the correlation with human judgments in precision tasks are significantly lower than those in recall tasks. This discrepancy arises from the inherent difficulty in evaluating precision compared to recall. 
Specifically, in the precision evaluation task, the evaluation focuses on assessing the correctness of model predictions. 
However, since the evaluated targets in the recall evaluation task are assumed to be correct, the evaluation primarily involves identifying potential correspondences. 
This phenomenon is also observed among human evaluators, as the correlation among humans is lower in the precision task compared to the recall task.


\subsubsection{Ablation of Adaptive Mechanism}
The proposed adaptive mechanism improves correlation with human in most cases, regardless of the evaluation agent. Besides, except for the deepseek series LLMs\footnote{The technical report of deepseek mentions that its LLM is sensitive to prompts \cite{deepseekai2025deepseekr1incentivizingreasoningcapability}. }, the adaptive mechanism significantly enhances the reliability of RAEE, as evidenced by a notable reduction in standard deviations across most evaluation results.
For example, the standard deviation for GPT-4o decreased by at least 0.36 (EAE precision) and by up to 4.46 (EAE recall) across all tasks.
This validates the effectiveness of the adaptive mechanism.


\subsubsection{Impart of Evaluation Agents Selection}
It is observed that the evaluation quality of RAEE depends on the capability of the evaluation agent. Recently, deepseek-r1 has been validated as the most comprehensive model across various tasks \cite{deepseekai2025deepseekr1incentivizingreasoningcapability}. 
In RAEE, deepseek-r1 delivers the overall strongest evaluation performance. Notably, it achieves a very strong correlation with human judgments in the EAE recall task, with a Percent agreement of 93.49\% and a Spearman correlation of 0.87, significantly outperforming other LLMs.

It is important to note that RAEE is a \textbf{trade-off} solution by balancing evaluation cost and reliability. 
When traditional methods fail to provide reliable semantic-level evaluation and pure human evaluation is costly, RAEE offers \textbf{a relatively reliable and affordable alternative}. As shown in Table~\ref{tab:raee_gpt4_vs_human}, the average time and financial cost of RAEE are significantly lower than those of pure human annotation. Furthermore, RAEE is not restricted by the choice of specific evaluation agents. 
With the improvement in the text evaluation capabilities of LLMs, the performance of RAEE will also improve, further reinforcing its reliability.

\begin{table}[h]
    \centering
    \small
    \begin{threeparttable}
        \begin{tabular}{l|cc}
        \toprule
           Evaluation Methods  & Time (hours) & Financial (\$) \\ \midrule
           RAEE + \textit{Deepseek-r1} & $\approx 5.85 $ & $\approx 5.44 $ \\
           RAEE + \textit{GPT-4o} & $\approx 0.20$ & $\approx 17.79$ \\
           Pure Human & $\approx 29.27$ & $\approx 255.13$ \\
        \bottomrule
        \end{tabular}
    \end{threeparttable}
    \caption{The average evaluation cost of human annotation and RAEE for an evaluated model on a dataset (approximate 1170 pieces of data on average). Assume RAEE evaluates with 10 parallel programs.} 
    \label{tab:raee_gpt4_vs_human}
\end{table}


\begin{table*}[h]    
    \centering
    \small
    \begin{threeparttable}
        \begin{tabular}{l|ccc|ccc|ccc|ccc}
            \toprule
            \multirow{2}{*}{\bf Model} & \multicolumn{3}{c|}{\bf ACE05} & \multicolumn{3}{c|}{\bf Genia2011} & \multicolumn{3}{c|}{\bf CASIE} & \multicolumn{3}{c}{\bf PHEE}\\
            ~ & p & r & f1 & p & r & f1 & p & r & f1 & p & r & f1 \\
            \midrule
            \multicolumn{13}{c}{\textit{Token-level Evaluation under \textbf{Exact Match}}} \\ 
            \midrule
            Tagprime-C & 62.69 & \cellcolor{red!30}74.60 & \cellcolor{red!10}68.13 & \cellcolor{red!10}75.16 & \cellcolor{red!10}71.91 & \cellcolor{red!10}73.50 & 68.29 & \cellcolor{red!10}71.85 & \cellcolor{red!10}70.02 & \cellcolor{red!10}72.56 & \cellcolor{red!10}70.39 & \cellcolor{red!30}71.46 \\
            OneIE & \cellcolor{red!30}70.94 & 70.69 & \cellcolor{red!30}70.82 & \cellcolor{red!30}75.52 & \cellcolor{red!30}72.30 & \cellcolor{red!30}73.87 & \cellcolor{red!10}69.68 & \cellcolor{red!30}72.32 & \cellcolor{red!30}70.98 & \cellcolor{red!30}74.86 & 67.10 & \cellcolor{red!10}70.77 \\
            Query\&Extract & \cellcolor{red!10}66.44 & 68.92 & 67.65 & 68.10 & 69.20 & 68.65 & 47.90 & 58.36 & 52.61 & 46.90 & 67.10 & 55.21 \\
            CEDAR & 60.30 & 71.76 & 65.53 & 72.19 & 64.88 & 68.34 & \cellcolor{red!30}70.05 & 68.06 & 69.04 & 69.88 & \cellcolor{red!30}71.49 & 70.68 \\
            DEGREE & 60.41 & \cellcolor{red!10}73.18 & 66.18 & 54.42 & 70.97 & 61.60 & 57.43 & 65.85 & 61.35 & 68.36 & 68.49 & 68.43 \\
            
            \midrule
            \multicolumn{13}{c}{\textit{Semantic-level Evaluation under \textbf{RAEE}}} \\ 
            \midrule
            Tagprime-C & 85.82 & \cellcolor{red!30}81.71 & \cellcolor{red!30}83.71 & \cellcolor{red!30}94.15 & \cellcolor{red!30}88.39 & \cellcolor{red!30}91.18 & 92.65 & \cellcolor{red!10}93.53 & \cellcolor{red!10}93.09 & \cellcolor{red!10}87.77 & 78.51 & \cellcolor{red!10}82.88 \\
            OneIE & \cellcolor{red!30}89.30 & 77.98 & \cellcolor{red!10}83.26 & \cellcolor{red!10}93.34 & 86.82 & \cellcolor{red!10}89.97 & \cellcolor{red!10}93.77 & \cellcolor{red!30}94.09 & \cellcolor{red!30}93.93 & \cellcolor{red!30}89.10 & 74.63 & 81.22 \\
            Query\&Extract & \cellcolor{red!10}87.16 & 76.38 & 81.41 & 91.88 & 85.90 & 88.78 & 81.75 & 91.88 & 86.52 & 66.83 & 75.52 & 70.91 \\
            CEDAR & 79.10 & 78.15 & 78.63 & 92.05 & 82.22 & 86.86 & \cellcolor{red!30}93.83 & 90.46 & 92.11 & 84.21 & \cellcolor{red!10}78.81 & 81.42 \\
            DEGREE & 82.70 & \cellcolor{red!10}81.53 & 82.11 & 88.95 & \cellcolor{red!10}87.84 & 88.39 & 92.71 & 90.54 & 91.61 & 86.07 & \cellcolor{red!30}80.20 & \cellcolor{red!30}83.03 \\
            
            \midrule
            
            \multirow{2}{*}{\bf Model} & \multicolumn{3}{c|}{\bf Fewevent}& \multicolumn{3}{c|}{\bf MEE-en}& \multicolumn{3}{c|}{\bf M2E2} & \multicolumn{3}{c}{\bf Average}\\
            ~ & p & r & f1 & p & r & f1 & p & r & f1 & p & r & f1 \\
            \midrule
            \multicolumn{13}{c}{\textit{Token-level Evaluation under \textbf{Exact Match}}} \\ 
            \midrule
            Tagprime-C & 63.62 & \cellcolor{red!30}66.00 & \cellcolor{red!10}64.79 & \cellcolor{red!10}79.27 & \cellcolor{red!10}82.18 & \cellcolor{red!30}80.70 & 46.35 & \cellcolor{red!10}62.07 & \cellcolor{red!10}53.07 & \cellcolor{red!10}66.85 & \cellcolor{red!30}71.28 & \cellcolor{red!10}68.81 \\
            OneIE & \cellcolor{red!30}68.38 & \cellcolor{red!10}62.65 & \cellcolor{red!30}65.39 & 78.22 & \cellcolor{red!30}82.52 & 80.31 & \cellcolor{red!30}52.10 & 50.00 & 51.03 & \cellcolor{red!30}69.96 & \cellcolor{red!10}68.23 & \cellcolor{red!30}69.02 \\
            Query\&Extract & \cellcolor{red!10}67.42 & 60.68 & 63.88 & 77.22 & \cellcolor{red!10}82.18 & 79.62 & \cellcolor{red!10}47.70 & \cellcolor{red!30}65.52 & \cellcolor{red!30}55.21 & 60.24 & 67.42 & 63.26 \\
            CEDAR & 44.18 & 62.26 & 51.68 & \cellcolor{red!30}81.71 & 79.37 & \cellcolor{red!10}80.52 & 45.54 & 52.87 & 48.94 & 63.41 & 67.24 & 64.96\\
            DEGREE & 60.33 & 59.78 & 60.05 & 78.06 & 78.68 & 78.37 & 43.46 & 59.20 & 50.12 & 60.35 & 68.02 & 63.73 \\
            
            \midrule
            \multicolumn{13}{c}{\textit{Semantic-level Evaluation under \textbf{RAEE}}} \\ 
            \midrule
            Tagprime-C & 88.24 & \cellcolor{red!30}73.87 & \cellcolor{red!30}80.42 & \cellcolor{red!10}93.42 & \cellcolor{red!10}86.99 & \cellcolor{red!30}90.09 & 94.42 & 64.94 & 76.96 & \cellcolor{red!10}90.93 & \cellcolor{red!30}81.13 & \cellcolor{red!30}85.48 \\
            OneIE & \cellcolor{red!30}90.03 & \cellcolor{red!10}69.85 & \cellcolor{red!10}78.67 & 93.21 & 86.65 & 89.81 & \cellcolor{red!30}96.41 & 53.45 & 68.77 & \cellcolor{red!30}92.17 & 77.64 & 83.66 \\
            Query\&Extract & \cellcolor{red!10}89.37 & 67.57 & 76.96 & 92.95 & \cellcolor{red!30}87.11 & \cellcolor{red!10}89.93 & 93.72 & \cellcolor{red!30}69.54 & \cellcolor{red!10}79.84 & 86.24 & 79.13 & 82.05 \\
            CEDAR & 61.83 & 69.30 & 65.35 & \cellcolor{red!30}94.40 & 83.27 & 88.48 & 89.11 & 57.47 & 69.88 & 84.93 & 77.10 & 80.39 \\
            DEGREE & 86.38 & 67.69 & 75.90 & 93.12 & 84.81 & 88.77 & \cellcolor{red!10}96.20 & \cellcolor{red!10}68.39 & \cellcolor{red!30}79.95 & 89.45 & \cellcolor{red!10}80.14 & \cellcolor{red!10}84.25 \\
            
            \bottomrule
        \end{tabular}
    \end{threeparttable}
    \caption{Evaluation under EM versus reassessment under RAEE on ED task. The values in the \textit{Average} column are derived from the mean of the values across 7 datasets. For each metric in each evaluation framework, we highlight the highest value using a dark shade, and the second-highest value using a light shade.}
    \label{tab:main_results_ed}
\end{table*}

\section{Reassessment Experiments}
The reliability of RAEE has been validated. In this section, we will apply RAEE to reassess the performance of existing EE models in mainstream datasets across various domains.
\subsection{Experimental Settings}
\paragraph{Tasks and Metrics}
ED and EAE tasks are evaluated separately.
Following \cite{lin-etal-2020-joint}, we focus on trigger classification (TC) and argument classification (AC) tasks.
The semantic-level f1-score for TC and AC is described in Section~\ref{sec:semantic_f1}.
Due to the time constraints, we select GPT-4o as the evaluation agent for RAEE.

\paragraph{Reassessed Models}
We reassess 8 state-of-the-art event extraction models using different frameworks. These models can be broadly categorized into two types: (1) Discriminative models that classify tokens into predefined categories, including Tagprime-C \cite{hsu-etal-2023-tagprime}, OneIE \cite{lin-etal-2020-joint}, Query\&Extract \cite{wang-etal-2022-query}, and CEDAR \cite{li-etal-2023-glen};
(2) Generative models that directly produce event-related outputs, including BartGen \cite{li-etal-2021-document}, PAIE \cite{ma-etal-2022-prompt}, DEGREE \cite{hsu-etal-2022-degree}, and AMPERE \cite{hsu-etal-2023-ampere}. Implementation details are shown in Appendix~\ref{sec:implementation_details}.

\paragraph{Event Datasets}
We evaluate the performance of event models on 10 datasets covering diverse domains, including news ACE05-en \cite{doddington-etal-2004-automatic} and RAMS \cite{ebner-etal-2020-multi}, general FewEvent \cite{10.1145/3336191.3371796} and GENEVA \cite{parekh-etal-2023-geneva}, Wikipedia MEE-en \cite{pouran-ben-veyseh-etal-2022-mee} and WikiEvents \cite{li-etal-2021-document}, biomedical Genia2011 \cite{kim-etal-2011-overview-genia}, multimedia M$^2$E$^2$ \cite{li-etal-2020-cross}, cybersecurity CASIE \cite{Satyapanich_Ferraro_Finin_2020}, and pharmacovigilance PHEE \cite{sun-etal-2022-phee}. 
Dataset statistics are provided in Appendix~\ref{sec:datasets}.

\begin{table*}[h]    
    \centering
    \small
    \begin{threeparttable}
        \begin{tabular}{l|ccc|ccc|ccc|ccc}
            \toprule
            \multirow{2}{*}{\bf Model} & \multicolumn{3}{c|}{\bf ACE05} & \multicolumn{3}{c|}{\bf Genia2011} & \multicolumn{3}{c|}{\bf CASIE} & \multicolumn{3}{c}{\bf PHEE} \\
            ~ & p & r & f1 & p & r & f1 & p & r & f1 & p & r & f1 \\
            \midrule
            \multicolumn{13}{c}{\textit{Token-level Evaluation under \textbf{Exact Match}}} \\ 
            \midrule
            Tagprime-C & \cellcolor{red!30}75.23 & \cellcolor{red!10}70.05 & \cellcolor{red!30}72.55 & \cellcolor{red!30}83.46 & \cellcolor{red!30}68.79 & \cellcolor{red!30}75.42 & \cellcolor{red!30}66.39 & \cellcolor{red!30}66.07 & \cellcolor{red!30}66.23 & \cellcolor{red!10}71.69 & 45.84 & \cellcolor{red!10}55.92 \\
            BartGen & 72.19 & 67.39 & 69.71 & 78.10 & 64.58 & 70.70 & 59.68 & 60.52 & 60.10 & 58.17 & 43.49 & 49.77 \\
            PAIE & 74.14 & \cellcolor{red!30}70.29 & \cellcolor{red!10}72.16 & \cellcolor{red!10}81.07 & \cellcolor{red!10}67.84 & \cellcolor{red!10}73.87 & \cellcolor{red!10}65.13 & \cellcolor{red!10}62.90 & \cellcolor{red!10}64.00 & \cellcolor{red!30}72.99 & \cellcolor{red!30}74.34 & \cellcolor{red!30}73.66 \\
            DEGREE & \cellcolor{red!10}74.30 & 67.39 & 70.68 & 70.08 & 57.95 & 63.44 & 55.29 & 51.32 & 53.23 & 61.42 & \cellcolor{red!10}47.63 & 53.65 \\
            AMPERE & 69.87 & 64.13 & 66.88 & 71.45 & 56.89 & 63.35 & 59.08 & 50.21 & 54.29 & 60.94 & 44.54 & 51.46 \\
            
            \midrule
            \multicolumn{13}{c}{\textit{Semantic-level Evaluation under \textbf{RAEE}}} \\ 
            \midrule
            Tagprime-C & \cellcolor{red!10}92.72 & 79.11  & \cellcolor{red!10}85.37 & \cellcolor{red!30}92.30 & 72.35 & 81.12 & \cellcolor{red!30}92.61 & \cellcolor{red!10}83.15 & \cellcolor{red!10}87.62 & \cellcolor{red!30}89.89 & 81.57 & \cellcolor{red!10}85.53 \\
            BartGen & 91.34 & \cellcolor{red!10}79.59 & 85.06 & \cellcolor{red!10}91.26 & \cellcolor{red!10}73.79 & \cellcolor{red!10}81.60 & 86.21 & 80.74 & 83.39 & 84.16 & 79.26 & 81.64 \\
            PAIE & 92.35 & \cellcolor{red!30}81.52 & \cellcolor{red!30}86.60 & 90.56 & \cellcolor{red!30}76.41 & \cellcolor{red!30}82.89 & \cellcolor{red!10}92.01 & \cellcolor{red!30}83.91 & \cellcolor{red!30}87.78 & \cellcolor{red!10}89.22 & \cellcolor{red!30}92.73 & \cellcolor{red!30}90.94 \\
            DEGREE & \cellcolor{red!30}93.35 & 77.29 & 84.57 & 86.87 & 68.14 & 76.37 & 88.16 & 71.82 & 79.16 & 86.75 & \cellcolor{red!10}82.37 & 84.50 \\
            AMPERE & 88.96 & 74.88 & 81.32 & 87.71 & 67.44 & 76.25 & 89.31 & 66.98 & 76.55 & 85.89 & 80.23 & 82.97 \\
            
            \midrule
            
            \multirow{2}{*}{\bf Model} & \multicolumn{3}{c|}{\bf WikiEvents}& \multicolumn{3}{c|}{\bf RAMS}& \multicolumn{3}{c|}{\bf GENEVA} & \multicolumn{3}{c}{\bf Average}\\
            ~ & p & r & f1 & p & r & f1 & p & r & f1 & p & r & f1 \\
            \midrule
            \multicolumn{13}{c}{\textit{Token-level Evaluation under \textbf{Exact Match}}} \\ 
            \midrule
            Tagprime-C & 66.80 & \cellcolor{red!10}63.69 & 65.21 & \cellcolor{red!10}50.34 & 45.02 & \cellcolor{red!10}47.53 & \cellcolor{red!30}78.25 & \cellcolor{red!30}78.99 & \cellcolor{red!30}78.62 & \cellcolor{red!30}70.31 & \cellcolor{red!10}62.64 & \cellcolor{red!10}65.93 \\
            BartGen & \cellcolor{red!10}70.24 & \cellcolor{red!10}63.69 & \cellcolor{red!10}66.80 & 48.70 & 43.37 & 45.88 & 64.20 & 63.69 & 63.94 & 64.47 & 58.10 & 60.99 \\
            PAIE & \cellcolor{red!30}70.71 & \cellcolor{red!30}65.63 & \cellcolor{red!30}68.08 & \cellcolor{red!30}51.94 & \cellcolor{red!30}51.06 & \cellcolor{red!30}51.50 & \cellcolor{red!10}70.43 & \cellcolor{red!10}68.41 & \cellcolor{red!10}69.41 & \cellcolor{red!10}69.49 & \cellcolor{red!30}65.78 & \cellcolor{red!30}67.52 \\
            DEGREE & 57.27 & 51.26 & 54.10 & 45.37 & 44.60 & 44.98 & 64.88 & 66.17 & 65.52 & 61.23 & 55.19 & 57.94 \\
            AMPERE & 60.42 & 50.10 & 54.78 & 47.87 & \cellcolor{red!10}46.63 & 47.24 & 65.41 & 66.62 & 66.01 & 62.15 & 54.16 & 57.71 \\
            
            \midrule
            \multicolumn{13}{c}{\textit{Semantic-level Evaluation under \textbf{RAEE}}} \\ 
            \midrule
            Tagprime-C & 93.32 & 76.36 & 83.99 & 81.31 & 70.14 & 75.31 & \cellcolor{red!30}93.65 & \cellcolor{red!30}92.75 & \cellcolor{red!30}93.20 & \cellcolor{red!30}90.83 & \cellcolor{red!10}79.35 & \cellcolor{red!10}84.59 \\
            BartGen & \cellcolor{red!30}93.99 & \cellcolor{red!10}76.94 & \cellcolor{red!10}84.61 & \cellcolor{red!10}81.49 & 68.72 & 74.56 & 90.34 & 83.36 & 86.71 & 88.40 & 77.49 & 82.51 \\
            PAIE & 92.74 & \cellcolor{red!30}79.84 & \cellcolor{red!30}85.81 & 81.23 & \cellcolor{red!30}74.89 & \cellcolor{red!30}77.93 & \cellcolor{red!10}93.25 & \cellcolor{red!10}88.77 & \cellcolor{red!10}90.96 & \cellcolor{red!10}90.19 & \cellcolor{red!30}82.58 & \cellcolor{red!30}86.13 \\
            DEGREE & 88.48 & 70.74 & 78.62 & 80.22 & \cellcolor{red!10}72.87 & 76.37 & 91.43 & 86.14 & 88.70 & 87.89 & 75.62 & 81.18 \\
            AMPERE & \cellcolor{red!10}93.61 & 67.64 & 78.53 & \cellcolor{red!30}81.70 & \cellcolor{red!10}72.87 & \cellcolor{red!10}77.03 & 91.56 & 86.64 & 89.03 & 88.39 & 73.81 & 80.24 \\
            
            \bottomrule
        \end{tabular}
    \end{threeparttable}
    \caption{Evaluation under EM versus reassessment under RAEE on EAE task. The values in the \textit{Average} column are derived from the mean of the values across 7 datasets. For each metric in each evaluation framework, we highlight the highest value using a dark shade, and the second-highest value using a light shade.}
    \label{tab:main_results_eae}
\end{table*}

\subsection{Main Results}
Table~\ref{tab:main_results_ed} and \ref{tab:main_results_eae} show both EM evaluations and RAEE reassessment results for ED and EAE tasks, respectively. 
Notably, across all models, RAEE reassessment shows a significant improvement in metrics compared to EM. 
On average, the precision of all models increased by more than 20.52\%, with recall improvement exceeding 9.85\% for ED and 16.71\% for EAE. 
It indicates that EM severely underestimates model performance by overlooking a substantial number of semantically correct predictions. Moreover, as shown in Table \ref{tab:category_delta_f1}, the evaluated models' performance gap between RAEE and EM is significantly larger for generative models compared to discriminative models, indicating that EM introduces a greater bias for generative models.

\begin{table}[h]
    \small
    \centering
    \begin{threeparttable}
    \begin{tabular}{c|l|c}
    \toprule
       \bf Task  & \bf Model Category & \bf Average $\Delta$ f1 \\
    \midrule
       \multirow{2}{*}{ED}  & Discriminative Models  & 16.38 \\
       ~ & Generative Models  & \textbf{20.52} \\
    \midrule
       \multirow{2}{*}{EAE}  & Discriminative Models  & 18.66 \\
       ~ & Generative Models  & \textbf{21.47} \\
       \bottomrule
    \end{tabular}
    \end{threeparttable}
    \caption{Average f1-score gap between EM and RAEE.}
    \label{tab:category_delta_f1}
\end{table}


Furthermore, in many datasets, model rankings change considerably when shifting EM to RAEE.
For instance, in the ED task, DEGREE ranks lower than OneIE and CEDAR across all datasets under EM, but outperforms both on average under RAEE. In the EAE task, rankings shift in ACE05, Genia2011 and CASIE datasets.
It suggests that EM distorts the true performance ranks of models, leading to misconceptions about model performance.

\subsection{Fine-grained Reason Analysis}

To explore the main reasons behind the misjudgments evaluated by EM and the primary failure modes in RAEE, we conduct a fine-grained analysis experiment, extending the reasons reported in \cite{han2024empiricalstudyinformationextraction, Chen_Qin_Jiang_Choi_2024}.
200 sampled pieces of data for both ED and EAE tasks are evaluated through multiple human experts.

\paragraph{Misjudgments under EM} 
As for instances evaluated as incorrect under EM but reassessed as correct under RAEE, we summarize 4 main categories of reasons: 
(1) \textbf{\textit{With CoreWord:}} Compared to the ground-truth, the predicted results contain the core words and only have missing or redundant words that do not affect the overall semantics; 
(2) \textbf{\textit{Co-reference:}} There is a relationship between the predicted result and a certain ground-truth refer to the same entity or concept; 
(3) \textbf{\textit{Unannotated Correct:}} While no match is found in ground-truth, the predicted results are still reassessed as correct; 
(4) \textit{Other non-representative reasons}.

\begin{figure}[h]
    \centering
    \includegraphics[width=0.9\linewidth]{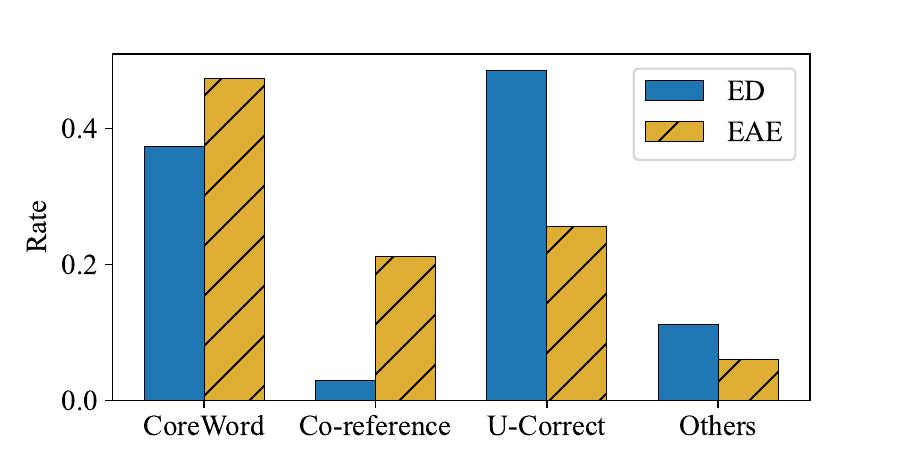}
    \caption{Distribution of the reasons that misjudges by EM evaluation method on ED and EAE tasks.}
    \label{fig:reason_analysis_correct}
\end{figure}

\begin{figure}[h]
    \centering
    \includegraphics[width=\linewidth]{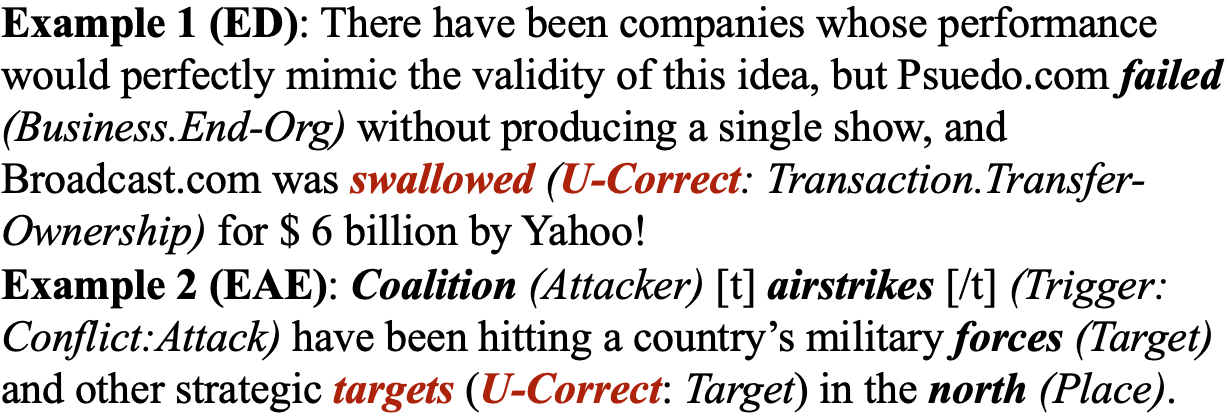}
    \caption{Two examples of \textit{Unannotated Correct} cases.}
    \label{fig:unannotated}
\end{figure}

As shown in Figure~\ref{fig:reason_analysis_correct}, the main misjudgments in EM are \textit{CoreWord} and \textit{U-Correct}. It indicates that current EE models tend to preserve core words while expanding or omitting some less relevant words, without changing the semantic meaning of extractions. 
The proportion of \textit{U-Correct} is similar to the 38.70\% reported in \cite{han2024empiricalstudyinformationextraction}. Two real examples are provided in Figure~\ref{fig:unannotated}.
This suggests that it may be necessary to introduce new rules, such as supplementary annotations, to enhance the comprehensiveness of some datasets.

\paragraph{Failure Modes under RAEE}
As for instances evaluated as incorrect under both EM and RAEE, we also summarize 4 kinds of reasons:
(1) \textbf{\textit{Without CoreWord:}} Compared to the ground-truth, the predicted results have missing core words that convey the primary semantics;
(2) \textbf{\textit{WrongType:}} Even though there is a semantic-level match of span in the ground-truth, the predicted results have an incorrect type of event or role;
(3) \textbf{\textit{Unannotated Incorrect:}} These instances are reassessed as incorrect and they are not included in the ground-truth;  
(4) \textit{Other non-representative reasons}.

\begin{figure}[h]
    \centering
    \includegraphics[width=0.9\linewidth]{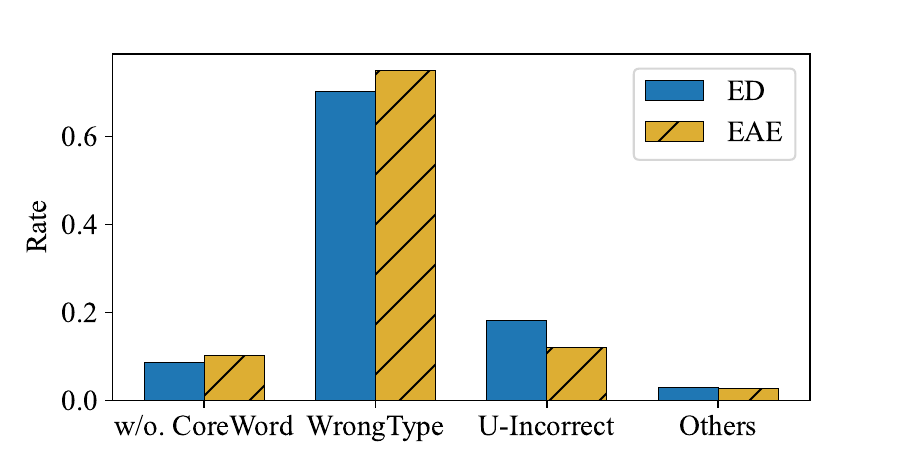}
    \caption{Distribution of failure modes under our proposed RAEE evaluation framework on ED and EAE.}
    \label{fig:reason_analysis_incorrect}
\end{figure}

As shown in Figure~\ref{fig:reason_analysis_incorrect}, the most frequent failure mode is \textit{WrongType}.
It indicates that while existing EE models correctly locate the spans, they often misclassify the types.
Future works should focus on enhancing models' ability to accurately understand and classify the types of extractions.

\section{LLMs Performance in EE Task}
Previous work has validated a significant performance gap between LLMs and state-of-the-art smaller fine-tuned models in the EE task under EM \cite{huang2024texteebenchmarkreevaluationreflections}.
In this section, we try to reassess LLMs performance in the EE task. Specifically, 6 advanced open-source and closed-source LLMs are reassessed: GPT-4o, GPT-3.5-turbo, Claude-3-Opus, Llama-3.1-70B, Qwen2-70B \cite{yang2024qwen2technicalreport} and Mixtral-8X7B \cite{jiang2024mixtralexperts}. The well-selected prompts for inference are shown in Appendix~\ref{sec:inference_prompts}.

Experiment results are shown in Table ~\ref{tab:llm_ed} and \ref{tab:llm_eae}.
It is found that LLMs generally underperform smaller fine-tuned models in both ED and EAE task, regardless of evaluation.
However, LLMs show much more improvements than smaller models when shifting EM to RAEE, indicating that LLMs' event extraction performance is greatly underestimated. 
Notably, f1-score improvement of LLMs on EAE task, over 51.92\%, is higher than that on ED task, over 30.74\%. 
This is mainly due to the fact that extractions in EAE task consist of more words (2.52 on average) than those in ED task (1.21 on average), making them more susceptible to misjudgment by EM.

\begin{table}[h]
    \centering
    \small
    \begin{threeparttable}
        \begin{tabular}{l|ccc|c}
        \toprule
            \multirow{2}{*}{\bf Model} & \multicolumn{3}{c|}{\bf RAEE} & \bf EM\\
            ~ & \bf p & \bf r & \bf f1 & \bf f1\\
        \midrule
           Tagprime-C  & \bf 93.75 & \bf 86.38 & \bf 89.91 & \bf 78.26 \\
           DEGREE & 89.31 & 85.91 & 87.58 & 72.48 \\
        \midrule 
           GPT-4o & \bf 79.08 & 53.05 & 63.50 & 32.76 \\
           \ \ \ \ \ \ - \textit{2 shots} & 77.66 & 48.35 & 59.60 & 33.17 \\
           \ \ \ \ \ \ - \textit{4 shots} & 78.77 & 49.76 & 60.99 & 28.23 \\
           \ \ \ \ \ \ - \textit{8 shots} & 74.41 & \bf 61.50 & \bf 67.34 & \bf 37.79 \\
           GPT-3.5 & 55.69 & 45.07 & 49.82 & 13.94 \\
           Claude-3-Opus & 76.31 & 40.84 & 53.21 & 13.69 \\
           Llama-3.1-70B & 72.60 & 54.46 & 62.23 & 16.66 \\
           Qwen2-70B-Chat & 61.53 & 55.39 & 58.30 & 14.47 \\    
           Mixtral-8x7B & 45.84 & 50.23 & 47.94 & 2.72 \\
        \bottomrule
        \end{tabular}
    \end{threeparttable}
    \caption{LLMs versus Top-2 state-of-the-art smaller fine-tuned models on ED task under RAEE evaluation.}
    \label{tab:llm_ed}
\end{table}

\begin{table}[h]
    \centering
    \small
    \begin{threeparttable}
        \begin{tabular}{l|ccc|c}
        \toprule
            \multirow{2}{*}{\bf Model} & \multicolumn{3}{c|}{\bf RAEE} & \bf EM\\
            ~ & \bf p & \bf r & \bf f1 & \bf f1\\
        \midrule
           PAIE & 89.90 & \bf 86.00 & \bf 87.91 & \bf 67.76 \\
           Tagprime-C & \bf 91.93 & 81.81 & 86.58 & 65.60 \\
        \midrule 
           GPT-4o & \bf 83.24 & \bf 73.19 & \bf 77.89 & 15.89\\
           \ \ \ \ \ \ - \textit{2 shots} & 82.49 & 69.56 & 75.48 & 16.36 \\
           \ \ \ \ \ \ - \textit{4 shots} & 81.75 & 69.02 & 74.85 & 16.38 \\
           \ \ \ \ \ \ - \textit{8 shots} & 80.02 & 69.92 & 74.63 & 15.57 \\
           GPT-3.5 & 62.87 & 64.13 & 63.49 & 10.26 \\
           Claude-3-Opus & 78.82 & 67.39 & 72.66 & \bf 20.22 \\
           Llama-3.1-70B & 60.62 & 69.56 & 64.78 & 12.86 \\
           Qwen2-70B-Chat & 79.95 & 68.11 & 73.56 & 16.99 \\
           Mixtral-8x7B & 62.17 & 65.94 & 64.00 & 11.76 \\
        \bottomrule
        \end{tabular}
    \end{threeparttable}
    \caption{LLMs versus Top-2 state-of-the-art smaller fine-tuned models on EAE task under RAEE evaluation.}
    \label{tab:llm_eae}
\end{table}

We also conduct few-shot experiments with the best-performing LLM, GPT-4o.
It is found that as the number of few-shot examples increases, the model's performance fluctuates on ED task and declines on EAE task. 
Our human annotations reveal that these results are heavily influenced by the quality of selected shots. 
Since the data to be extracted is sampled from datasets across diverse domains, using few-shot examples from only a limited set of domains is more likely to mislead the model.

\section{Conclusion and Future Works}

In this paper, we introduce RAEE, a reliable semantic-level evaluation framework for event extraction, which moves beyond the token-level exact match evaluation.
Extensive experiments validate RAEE's strong correlation with human judgments, confirming its reliability.
We apply RAEE to reassess the performance of numerous prior works and advanced LLMs in EE.
Extensive experiments indicate that both prior works and LLMs are significantly underestimated. 
A further fine-grained reason analysis reveals several intriguing phenomena worth further exploration.

In the future, we will extend the evaluation of model performance with RAEE from closed domains to the open domain.

\section*{Limitations}

\paragraph{LLM Judgments \textit{v.s.} Human Judgments}
In this paper, the advanced LLM GPT-4o is used to automatically judge predictions from event extraction models.
However, it should be noted that LLMs cannot always perfectly align with human judgments, which is the most reliable evaluation.
For example, GPT-4o achieves comparable agreement and correlations with human average performance in ED and EAE evaluations. 
While we have made efforts to improve the relevance between LLM and human assessment through prompt design, this gap still exists.
In the future, we would like to discover other strategies to further improve the reliability of LLM-as-a-Judge, such as multi-agent debate \cite{chan2023chatevalbetterllmbasedevaluators}.

\paragraph{Sensitivity of LLMs to Prompts}
In our preliminary study, it is found that the evaluation agent LLMs are sensitive to prompts. In the meta-evaluation experiments, using different prompts significantly affects LLMs' correlation with human judgments. To mitigate this problem, we provide well-designed prompts that performs well across all LLMs in the meta-evaluation experiments, which can be found in Appendix~\ref{sec:reassess_prompts}.

\paragraph{Computational Cost of Evaluation}
Unlike existing EM evaluation, our proposed RAEE relies on advanced LLMs to evaluate the quality of predicted events, leading to much higher costs.
The average cost of evaluating one model on a dataset is between 2-5\$, which is smaller than established benchmarks like AlpacaEval \cite{alpaca_eval}.

\paragraph{Insufficient Evaluation for LLMs}
The number of the test set in existing event extract datasets contains thousands of samples, leading to a huge cost for LLMs to do inference experiments. Following previous works \cite{huang2024texteebenchmarkreevaluationreflections}, we sample a subset of the test set in existing datasets and prompt LLMs to handle the event extraction task.
However, we wish to emphasize that the subset consists of diverse samples from 10 event extraction datasets. Therefore, the conclusions of our experimental results are reliable and robust.

\bibliography{main}

\begin{thebibliography}{52}
\providecommand{\natexlab}[1]{#1}

\bibitem[{Chan et~al.(2023)Chan, Chen, Su, Yu, Xue, Zhang, Fu, and Liu}]{chan2023chatevalbetterllmbasedevaluators}
Chi-Min Chan, Weize Chen, Yusheng Su, Jianxuan Yu, Wei Xue, Shanghang Zhang, Jie Fu, and Zhiyuan Liu. 2023.
\newblock \href {https://arxiv.org/abs/2308.07201} {Chateval: Towards better llm-based evaluators through multi-agent debate}.
\newblock \emph{Preprint}, arXiv:2308.07201.

\bibitem[{Chen et~al.(2023)Chen, Xiao, Zhang, Luo, Lian, and Liu}]{bge_m3}
Jianlv Chen, Shitao Xiao, Peitian Zhang, Kun Luo, Defu Lian, and Zheng Liu. 2023.
\newblock \href {https://arxiv.org/abs/2309.07597} {Bge m3-embedding: Multi-lingual, multi-functionality, multi-granularity text embeddings through self-knowledge distillation}.
\newblock \emph{Preprint}, arXiv:2309.07597.

\bibitem[{Chen et~al.(2024{\natexlab{a}})Chen, Qin, Jiang, and Choi}]{Chen_Qin_Jiang_Choi_2024}
Ruirui Chen, Chengwei Qin, Weifeng Jiang, and Dongkyu Choi. 2024{\natexlab{a}}.
\newblock \href {https://doi.org/10.1609/aaai.v38i16.29730} {Is a large language model a good annotator for event extraction?}
\newblock \emph{Proceedings of the AAAI Conference on Artificial Intelligence}, 38(16):17772--17780.

\bibitem[{Chen et~al.(2024{\natexlab{b}})Chen, Du, Zhang, Liu, Liu, Zheng, Zhuo, Zhang, Lin, Chen, and Zhao}]{chen2024tevalevaluatingtoolutilization}
Zehui Chen, Weihua Du, Wenwei Zhang, Kuikun Liu, Jiangning Liu, Miao Zheng, Jingming Zhuo, Songyang Zhang, Dahua Lin, Kai Chen, and Feng Zhao. 2024{\natexlab{b}}.
\newblock \href {https://arxiv.org/abs/2312.14033} {T-eval: Evaluating the tool utilization capability of large language models step by step}.
\newblock \emph{Preprint}, arXiv:2312.14033.

\bibitem[{DeepSeek-AI et~al.(2025)DeepSeek-AI, Guo, Yang, Zhang, Song, Zhang, Xu, Zhu, Ma, Wang, Bi, Zhang, Yu, Wu, Wu, Gou, Shao, Li, Gao, Liu, Xue, Wang, Wu, Feng, Lu, Zhao, Deng, Zhang, Ruan, Dai, Chen, Ji, Li, Lin, Dai, Luo, Hao, Chen, Li, Zhang, Bao, Xu, Wang, Ding, Xin, Gao, Qu, Li, Guo, Li, Wang, Chen, Yuan, Qiu, Li, Cai, Ni, Liang, Chen, Dong, Hu, Gao, Guan, Huang, Yu, Wang, Zhang, Zhao, Wang, Zhang, Xu, Xia, Zhang, Zhang, Tang, Li, Wang, Li, Tian, Huang, Zhang, Wang, Chen, Du, Ge, Zhang, Pan, Wang, Chen, Jin, Chen, Lu, Zhou, Chen, Ye, Wang, Yu, Zhou, Pan, Li, Zhou, Wu, Ye, Yun, Pei, Sun, Wang, Zeng, Zhao, Liu, Liang, Gao, Yu, Zhang, Xiao, An, Liu, Wang, Chen, Nie, Cheng, Liu, Xie, Liu, Yang, Li, Su, Lin, Li, Jin, Shen, Chen, Sun, Wang, Song, Zhou, Wang, Shan, Li, Wang, Wei, Zhang, Xu, Li, Zhao, Sun, Wang, Yu, Zhang, Shi, Xiong, He, Piao, Wang, Tan, Ma, Liu, Guo, Ou, Wang, Gong, Zou, He, Xiong, Luo, You, Liu, Zhou, Zhu, Xu, Huang, Li, Zheng, Zhu, Ma, Tang, Zha, Yan, Ren, Ren, Sha, Fu, Xu, Xie, Zhang,
  Hao, Ma, Yan, Wu, Gu, Zhu, Liu, Li, Xie, Song, Pan, Huang, Xu, Zhang, and Zhang}]{deepseekai2025deepseekr1incentivizingreasoningcapability}
DeepSeek-AI, Daya Guo, Dejian Yang, Haowei Zhang, Junxiao Song, Ruoyu Zhang, Runxin Xu, Qihao Zhu, Shirong Ma, Peiyi Wang, Xiao Bi, Xiaokang Zhang, Xingkai Yu, Yu~Wu, Z.~F. Wu, Zhibin Gou, Zhihong Shao, Zhuoshu Li, Ziyi Gao, Aixin Liu, Bing Xue, Bingxuan Wang, Bochao Wu, Bei Feng, Chengda Lu, Chenggang Zhao, Chengqi Deng, Chenyu Zhang, Chong Ruan, Damai Dai, Deli Chen, Dongjie Ji, Erhang Li, Fangyun Lin, Fucong Dai, Fuli Luo, Guangbo Hao, Guanting Chen, Guowei Li, H.~Zhang, Han Bao, Hanwei Xu, Haocheng Wang, Honghui Ding, Huajian Xin, Huazuo Gao, Hui Qu, Hui Li, Jianzhong Guo, Jiashi Li, Jiawei Wang, Jingchang Chen, Jingyang Yuan, Junjie Qiu, Junlong Li, J.~L. Cai, Jiaqi Ni, Jian Liang, Jin Chen, Kai Dong, Kai Hu, Kaige Gao, Kang Guan, Kexin Huang, Kuai Yu, Lean Wang, Lecong Zhang, Liang Zhao, Litong Wang, Liyue Zhang, Lei Xu, Leyi Xia, Mingchuan Zhang, Minghua Zhang, Minghui Tang, Meng Li, Miaojun Wang, Mingming Li, Ning Tian, Panpan Huang, Peng Zhang, Qiancheng Wang, Qinyu Chen, Qiushi Du, Ruiqi Ge, Ruisong
  Zhang, Ruizhe Pan, Runji Wang, R.~J. Chen, R.~L. Jin, Ruyi Chen, Shanghao Lu, Shangyan Zhou, Shanhuang Chen, Shengfeng Ye, Shiyu Wang, Shuiping Yu, Shunfeng Zhou, Shuting Pan, S.~S. Li, Shuang Zhou, Shaoqing Wu, Shengfeng Ye, Tao Yun, Tian Pei, Tianyu Sun, T.~Wang, Wangding Zeng, Wanjia Zhao, Wen Liu, Wenfeng Liang, Wenjun Gao, Wenqin Yu, Wentao Zhang, W.~L. Xiao, Wei An, Xiaodong Liu, Xiaohan Wang, Xiaokang Chen, Xiaotao Nie, Xin Cheng, Xin Liu, Xin Xie, Xingchao Liu, Xinyu Yang, Xinyuan Li, Xuecheng Su, Xuheng Lin, X.~Q. Li, Xiangyue Jin, Xiaojin Shen, Xiaosha Chen, Xiaowen Sun, Xiaoxiang Wang, Xinnan Song, Xinyi Zhou, Xianzu Wang, Xinxia Shan, Y.~K. Li, Y.~Q. Wang, Y.~X. Wei, Yang Zhang, Yanhong Xu, Yao Li, Yao Zhao, Yaofeng Sun, Yaohui Wang, Yi~Yu, Yichao Zhang, Yifan Shi, Yiliang Xiong, Ying He, Yishi Piao, Yisong Wang, Yixuan Tan, Yiyang Ma, Yiyuan Liu, Yongqiang Guo, Yuan Ou, Yuduan Wang, Yue Gong, Yuheng Zou, Yujia He, Yunfan Xiong, Yuxiang Luo, Yuxiang You, Yuxuan Liu, Yuyang Zhou, Y.~X. Zhu,
  Yanhong Xu, Yanping Huang, Yaohui Li, Yi~Zheng, Yuchen Zhu, Yunxian Ma, Ying Tang, Yukun Zha, Yuting Yan, Z.~Z. Ren, Zehui Ren, Zhangli Sha, Zhe Fu, Zhean Xu, Zhenda Xie, Zhengyan Zhang, Zhewen Hao, Zhicheng Ma, Zhigang Yan, Zhiyu Wu, Zihui Gu, Zijia Zhu, Zijun Liu, Zilin Li, Ziwei Xie, Ziyang Song, Zizheng Pan, Zhen Huang, Zhipeng Xu, Zhongyu Zhang, and Zhen Zhang. 2025.
\newblock \href {https://arxiv.org/abs/2501.12948} {Deepseek-r1: Incentivizing reasoning capability in llms via reinforcement learning}.
\newblock \emph{Preprint}, arXiv:2501.12948.

\bibitem[{Deng et~al.(2020)Deng, Zhang, Kang, Zhang, Zhang, and Chen}]{10.1145/3336191.3371796}
Shumin Deng, Ningyu Zhang, Jiaojian Kang, Yichi Zhang, Wei Zhang, and Huajun Chen. 2020.
\newblock \href {https://doi.org/10.1145/3336191.3371796} {Meta-learning with dynamic-memory-based prototypical network for few-shot event detection}.
\newblock In \emph{Proceedings of the 13th International Conference on Web Search and Data Mining}, WSDM '20, page 151–159, New York, NY, USA. Association for Computing Machinery.

\bibitem[{Doddington et~al.(2004)Doddington, Mitchell, Przybocki, Ramshaw, Strassel, and Weischedel}]{doddington-etal-2004-automatic}
George Doddington, Alexis Mitchell, Mark Przybocki, Lance Ramshaw, Stephanie Strassel, and Ralph Weischedel. 2004.
\newblock \href {http://www.lrec-conf.org/proceedings/lrec2004/pdf/5.pdf} {The automatic content extraction ({ACE}) program {--} tasks, data, and evaluation}.
\newblock In \emph{Proceedings of the Fourth International Conference on Language Resources and Evaluation ({LREC}{'}04)}, Lisbon, Portugal. European Language Resources Association (ELRA).

\bibitem[{Ebner et~al.(2020)Ebner, Xia, Culkin, Rawlins, and Van~Durme}]{ebner-etal-2020-multi}
Seth Ebner, Patrick Xia, Ryan Culkin, Kyle Rawlins, and Benjamin Van~Durme. 2020.
\newblock \href {https://doi.org/10.18653/v1/2020.acl-main.718} {Multi-sentence argument linking}.
\newblock In \emph{Proceedings of the 58th Annual Meeting of the Association for Computational Linguistics}, pages 8057--8077, Online. Association for Computational Linguistics.

\bibitem[{Fu et~al.(2023)Fu, Ng, Jiang, and Liu}]{fu2023gptscore}
Jinlan Fu, See-Kiong Ng, Zhengbao Jiang, and Pengfei Liu. 2023.
\newblock Gptscore: Evaluate as you desire.
\newblock \emph{arXiv preprint arXiv:2302.04166}.

\bibitem[{Gao et~al.(2023)Gao, Zhao, Yu, and Xu}]{gao2023exploringfeasibilitychatgptevent}
Jun Gao, Huan Zhao, Changlong Yu, and Ruifeng Xu. 2023.
\newblock \href {https://arxiv.org/abs/2303.03836} {Exploring the feasibility of chatgpt for event extraction}.
\newblock \emph{Preprint}, arXiv:2303.03836.

\bibitem[{Grishman(2015)}]{7243219}
Ralph Grishman. 2015.
\newblock \href {https://doi.org/10.1109/MIS.2015.68} {Information extraction}.
\newblock \emph{IEEE Intelligent Systems}, 30(5):8--15.

\bibitem[{Han et~al.(2024)Han, Yang, Peng, Tiwari, Wan, Liu, and Wang}]{han2024empiricalstudyinformationextraction}
Ridong Han, Chaohao Yang, Tao Peng, Prayag Tiwari, Xiang Wan, Lu~Liu, and Benyou Wang. 2024.
\newblock \href {https://arxiv.org/abs/2305.14450} {An empirical study on information extraction using large language models}.
\newblock \emph{Preprint}, arXiv:2305.14450.

\bibitem[{Han et~al.(2021)Han, Hsu, Sun, Baylon, Ning, Roth, and Peng}]{han-etal-2021-ester}
Rujun Han, I-Hung Hsu, Jiao Sun, Julia Baylon, Qiang Ning, Dan Roth, and Nanyun Peng. 2021.
\newblock \href {https://doi.org/10.18653/v1/2021.emnlp-main.597} {{ESTER}: A machine reading comprehension dataset for reasoning about event semantic relations}.
\newblock In \emph{Proceedings of the 2021 Conference on Empirical Methods in Natural Language Processing}, pages 7543--7559, Online and Punta Cana, Dominican Republic. Association for Computational Linguistics.

\bibitem[{Hsu et~al.(2022)Hsu, Huang, Boschee, Miller, Natarajan, Chang, and Peng}]{hsu-etal-2022-degree}
I-Hung Hsu, Kuan-Hao Huang, Elizabeth Boschee, Scott Miller, Prem Natarajan, Kai-Wei Chang, and Nanyun Peng. 2022.
\newblock \href {https://doi.org/10.18653/v1/2022.naacl-main.138} {{DEGREE}: A data-efficient generation-based event extraction model}.
\newblock In \emph{Proceedings of the 2022 Conference of the North American Chapter of the Association for Computational Linguistics: Human Language Technologies}, pages 1890--1908, Seattle, United States. Association for Computational Linguistics.

\bibitem[{Hsu et~al.(2023{\natexlab{a}})Hsu, Huang, Zhang, Cheng, Natarajan, Chang, and Peng}]{hsu-etal-2023-tagprime}
I-Hung Hsu, Kuan-Hao Huang, Shuning Zhang, Wenxin Cheng, Prem Natarajan, Kai-Wei Chang, and Nanyun Peng. 2023{\natexlab{a}}.
\newblock \href {https://doi.org/10.18653/v1/2023.acl-long.723} {{TAGPRIME}: A unified framework for relational structure extraction}.
\newblock In \emph{Proceedings of the 61st Annual Meeting of the Association for Computational Linguistics (Volume 1: Long Papers)}, pages 12917--12932, Toronto, Canada. Association for Computational Linguistics.

\bibitem[{Hsu et~al.(2023{\natexlab{b}})Hsu, Xie, Huang, Natarajan, and Peng}]{hsu-etal-2023-ampere}
I-Hung Hsu, Zhiyu Xie, Kuan-Hao Huang, Prem Natarajan, and Nanyun Peng. 2023{\natexlab{b}}.
\newblock \href {https://doi.org/10.18653/v1/2023.acl-long.615} {{AMPERE}: {AMR}-aware prefix for generation-based event argument extraction model}.
\newblock In \emph{Proceedings of the 61st Annual Meeting of the Association for Computational Linguistics (Volume 1: Long Papers)}, pages 10976--10993, Toronto, Canada. Association for Computational Linguistics.

\bibitem[{Huang et~al.(2024)Huang, Hsu, Parekh, Xie, Zhang, Natarajan, Chang, Peng, and Ji}]{huang2024texteebenchmarkreevaluationreflections}
Kuan-Hao Huang, I-Hung Hsu, Tanmay Parekh, Zhiyu Xie, Zixuan Zhang, Premkumar Natarajan, Kai-Wei Chang, Nanyun Peng, and Heng Ji. 2024.
\newblock \href {https://arxiv.org/abs/2311.09562} {Textee: Benchmark, reevaluation, reflections, and future challenges in event extraction}.
\newblock \emph{Preprint}, arXiv:2311.09562.

\bibitem[{Jiang et~al.(2024)Jiang, Sablayrolles, Roux, Mensch, Savary, Bamford, Chaplot, de~las Casas, Hanna, Bressand, Lengyel, Bour, Lample, Lavaud, Saulnier, Lachaux, Stock, Subramanian, Yang, Antoniak, Scao, Gervet, Lavril, Wang, Lacroix, and Sayed}]{jiang2024mixtralexperts}
Albert~Q. Jiang, Alexandre Sablayrolles, Antoine Roux, Arthur Mensch, Blanche Savary, Chris Bamford, Devendra~Singh Chaplot, Diego de~las Casas, Emma~Bou Hanna, Florian Bressand, Gianna Lengyel, Guillaume Bour, Guillaume Lample, Lélio~Renard Lavaud, Lucile Saulnier, Marie-Anne Lachaux, Pierre Stock, Sandeep Subramanian, Sophia Yang, Szymon Antoniak, Teven~Le Scao, Théophile Gervet, Thibaut Lavril, Thomas Wang, Timothée Lacroix, and William~El Sayed. 2024.
\newblock \href {https://arxiv.org/abs/2401.04088} {Mixtral of experts}.
\newblock \emph{Preprint}, arXiv:2401.04088.

\bibitem[{Kim et~al.(2011)Kim, Wang, Takagi, and Yonezawa}]{kim-etal-2011-overview-genia}
Jin-Dong Kim, Yue Wang, Toshihisa Takagi, and Akinori Yonezawa. 2011.
\newblock \href {https://aclanthology.org/W11-1802} {Overview of {G}enia event task in {B}io{NLP} shared task 2011}.
\newblock In \emph{Proceedings of {B}io{NLP} Shared Task 2011 Workshop}, pages 7--15, Portland, Oregon, USA. Association for Computational Linguistics.

\bibitem[{Kim et~al.(2023)Kim, Shin, Cho, Jang, Longpre, Lee, Yun, Shin, Kim, Thorne et~al.}]{kim2023prometheus}
Seungone Kim, Jamin Shin, Yejin Cho, Joel Jang, Shayne Longpre, Hwaran Lee, Sangdoo Yun, Seongjin Shin, Sungdong Kim, James Thorne, et~al. 2023.
\newblock Prometheus: Inducing fine-grained evaluation capability in language models.
\newblock \emph{arXiv preprint arXiv:2310.08491}.

\bibitem[{Lan et~al.(2020)Lan, Mao, Wei, Gao, and Huang}]{lan2020pone}
Tian Lan, Xian-Ling Mao, Wei Wei, Xiaoyan Gao, and Heyan Huang. 2020.
\newblock Pone: A novel automatic evaluation metric for open-domain generative dialogue systems.
\newblock \emph{ACM Transactions on Information Systems (TOIS)}, 39(1):1--37.

\bibitem[{Lan et~al.(2024)Lan, Zhang, Xu, Huang, Lin, Chen, and Mao}]{lan2024criticbench}
Tian Lan, Wenwei Zhang, Chen Xu, Heyan Huang, Dahua Lin, Kai Chen, and Xian-ling Mao. 2024.
\newblock Criticbench: Evaluating large language models as critic.
\newblock \emph{arXiv preprint arXiv:2402.13764}.

\bibitem[{Lewis et~al.(2019)Lewis, Liu, Goyal, Ghazvininejad, Mohamed, Levy, Stoyanov, and Zettlemoyer}]{lewis2019bartdenoisingsequencetosequencepretraining}
Mike Lewis, Yinhan Liu, Naman Goyal, Marjan Ghazvininejad, Abdelrahman Mohamed, Omer Levy, Ves Stoyanov, and Luke Zettlemoyer. 2019.
\newblock \href {https://arxiv.org/abs/1910.13461} {Bart: Denoising sequence-to-sequence pre-training for natural language generation, translation, and comprehension}.
\newblock \emph{Preprint}, arXiv:1910.13461.

\bibitem[{Li et~al.(2020)Li, Zareian, Zeng, Whitehead, Lu, Ji, and Chang}]{li-etal-2020-cross}
Manling Li, Alireza Zareian, Qi~Zeng, Spencer Whitehead, Di~Lu, Heng Ji, and Shih-Fu Chang. 2020.
\newblock \href {https://doi.org/10.18653/v1/2020.acl-main.230} {Cross-media structured common space for multimedia event extraction}.
\newblock In \emph{Proceedings of the 58th Annual Meeting of the Association for Computational Linguistics}, pages 2557--2568, Online. Association for Computational Linguistics.

\bibitem[{Li et~al.(2021)Li, Ji, and Han}]{li-etal-2021-document}
Sha Li, Heng Ji, and Jiawei Han. 2021.
\newblock \href {https://doi.org/10.18653/v1/2021.naacl-main.69} {Document-level event argument extraction by conditional generation}.
\newblock In \emph{Proceedings of the 2021 Conference of the North American Chapter of the Association for Computational Linguistics: Human Language Technologies}, pages 894--908, Online. Association for Computational Linguistics.

\bibitem[{Li et~al.(2023{\natexlab{a}})Li, Zhan, Conger, Palmer, Ji, and Han}]{li-etal-2023-glen}
Sha Li, Qiusi Zhan, Kathryn Conger, Martha Palmer, Heng Ji, and Jiawei Han. 2023{\natexlab{a}}.
\newblock \href {https://doi.org/10.18653/v1/2023.emnlp-main.170} {{GLEN}: General-purpose event detection for thousands of types}.
\newblock In \emph{Proceedings of the 2023 Conference on Empirical Methods in Natural Language Processing}, pages 2823--2838, Singapore. Association for Computational Linguistics.

\bibitem[{Li et~al.(2023{\natexlab{b}})Li, Zhang, Dubois, Taori, Gulrajani, Guestrin, Liang, and Hashimoto}]{alpaca_eval}
Xuechen Li, Tianyi Zhang, Yann Dubois, Rohan Taori, Ishaan Gulrajani, Carlos Guestrin, Percy Liang, and Tatsunori~B. Hashimoto. 2023{\natexlab{b}}.
\newblock Alpacaeval: An automatic evaluator of instruction-following models.
\newblock \url{https://github.com/tatsu-lab/alpaca_eval}.

\bibitem[{Lin et~al.(2020)Lin, Ji, Huang, and Wu}]{lin-etal-2020-joint}
Ying Lin, Heng Ji, Fei Huang, and Lingfei Wu. 2020.
\newblock \href {https://doi.org/10.18653/v1/2020.acl-main.713} {A joint neural model for information extraction with global features}.
\newblock In \emph{Proceedings of the 58th Annual Meeting of the Association for Computational Linguistics}, pages 7999--8009, Online. Association for Computational Linguistics.

\bibitem[{Lin et~al.(2024)Lin, Gou, Liang, Luo, Liu, and Yang}]{lin2024criticbench}
Zicheng Lin, Zhibin Gou, Tian Liang, Ruilin Luo, Haowei Liu, and Yujiu Yang. 2024.
\newblock Criticbench: Benchmarking llms for critique-correct reasoning.
\newblock \emph{arXiv preprint arXiv:2402.14809}.

\bibitem[{Liu et~al.(2021)Liu, Zheng, Chang, and Sui}]{Liu2021TowardsFI}
Tianyu Liu, Xin Zheng, Baobao Chang, and Zhifang Sui. 2021.
\newblock \href {https://api.semanticscholar.org/CorpusID:231942490} {Towards faithfulness in open domain table-to-text generation from an entity-centric view}.
\newblock In \emph{AAAI Conference on Artificial Intelligence}.

\bibitem[{Liu et~al.(2023{\natexlab{a}})Liu, Yu, Zhang, Xu, Lei, Lai, Gu, Ding, Men, Yang, Zhang, Deng, Zeng, Du, Zhang, Shen, Zhang, Su, Sun, Huang, Dong, and Tang}]{liu2023agentbenchevaluatingllmsagents}
Xiao Liu, Hao Yu, Hanchen Zhang, Yifan Xu, Xuanyu Lei, Hanyu Lai, Yu~Gu, Hangliang Ding, Kaiwen Men, Kejuan Yang, Shudan Zhang, Xiang Deng, Aohan Zeng, Zhengxiao Du, Chenhui Zhang, Sheng Shen, Tianjun Zhang, Yu~Su, Huan Sun, Minlie Huang, Yuxiao Dong, and Jie Tang. 2023{\natexlab{a}}.
\newblock \href {https://arxiv.org/abs/2308.03688} {Agentbench: Evaluating llms as agents}.
\newblock \emph{Preprint}, arXiv:2308.03688.

\bibitem[{Liu et~al.(2023{\natexlab{b}})Liu, Iter, Xu, Wang, Xu, and Zhu}]{liu2023g}
Yang Liu, Dan Iter, Yichong Xu, Shuohang Wang, Ruochen Xu, and Chenguang Zhu. 2023{\natexlab{b}}.
\newblock G-eval: Nlg evaluation using gpt-4 with better human alignment.
\newblock \emph{arXiv preprint arXiv:2303.16634}.

\bibitem[{Liu et~al.(2019)Liu, Ott, Goyal, Du, Joshi, Chen, Levy, Lewis, Zettlemoyer, and Stoyanov}]{liu2019robertarobustlyoptimizedbert}
Yinhan Liu, Myle Ott, Naman Goyal, Jingfei Du, Mandar Joshi, Danqi Chen, Omer Levy, Mike Lewis, Luke Zettlemoyer, and Veselin Stoyanov. 2019.
\newblock \href {https://arxiv.org/abs/1907.11692} {Roberta: A robustly optimized bert pretraining approach}.
\newblock \emph{Preprint}, arXiv:1907.11692.

\bibitem[{Lu et~al.(2022)Lu, Liu, Dai, Xiao, Lin, Han, Sun, and Wu}]{lu-etal-2022-unified}
Yaojie Lu, Qing Liu, Dai Dai, Xinyan Xiao, Hongyu Lin, Xianpei Han, Le~Sun, and Hua Wu. 2022.
\newblock \href {https://doi.org/10.18653/v1/2022.acl-long.395} {Unified structure generation for universal information extraction}.
\newblock In \emph{Proceedings of the 60th Annual Meeting of the Association for Computational Linguistics (Volume 1: Long Papers)}, pages 5755--5772, Dublin, Ireland. Association for Computational Linguistics.

\bibitem[{Luo et~al.(2023)Luo, Lin, Liu, Shu, Zhu, Shang, and Meng}]{luo2023critique}
Liangchen Luo, Zi~Lin, Yinxiao Liu, Lei Shu, Yun Zhu, Jingbo Shang, and Lei Meng. 2023.
\newblock Critique ability of large language models.
\newblock \emph{arXiv preprint arXiv:2310.04815}.

\bibitem[{Ma et~al.(2022)Ma, Wang, Cao, Li, Chen, Wang, and Shao}]{ma-etal-2022-prompt}
Yubo Ma, Zehao Wang, Yixin Cao, Mukai Li, Meiqi Chen, Kun Wang, and Jing Shao. 2022.
\newblock \href {https://doi.org/10.18653/v1/2022.acl-long.466} {{P}rompt for extraction? {PAIE}: {P}rompting argument interaction for event argument extraction}.
\newblock In \emph{Proceedings of the 60th Annual Meeting of the Association for Computational Linguistics (Volume 1: Long Papers)}, pages 6759--6774, Dublin, Ireland. Association for Computational Linguistics.

\bibitem[{Parekh et~al.(2023)Parekh, Hsu, Huang, Chang, and Peng}]{parekh-etal-2023-geneva}
Tanmay Parekh, I-Hung Hsu, Kuan-Hao Huang, Kai-Wei Chang, and Nanyun Peng. 2023.
\newblock \href {https://doi.org/10.18653/v1/2023.acl-long.203} {{GENEVA}: Benchmarking generalizability for event argument extraction with hundreds of event types and argument roles}.
\newblock In \emph{Proceedings of the 61st Annual Meeting of the Association for Computational Linguistics (Volume 1: Long Papers)}, pages 3664--3686, Toronto, Canada. Association for Computational Linguistics.

\bibitem[{Park et~al.(2023)Park, O'Brien, Cai, Morris, Liang, and Bernstein}]{park2023generativeagentsinteractivesimulacra}
Joon~Sung Park, Joseph~C. O'Brien, Carrie~J. Cai, Meredith~Ringel Morris, Percy Liang, and Michael~S. Bernstein. 2023.
\newblock \href {https://arxiv.org/abs/2304.03442} {Generative agents: Interactive simulacra of human behavior}.
\newblock \emph{Preprint}, arXiv:2304.03442.

\bibitem[{Peng et~al.(2023{\natexlab{a}})Peng, Wang, Yao, Wang, Zhu, Zeng, Hou, and Li}]{peng-etal-2023-omnievent}
Hao Peng, Xiaozhi Wang, Feng Yao, Zimu Wang, Chuzhao Zhu, Kaisheng Zeng, Lei Hou, and Juanzi Li. 2023{\natexlab{a}}.
\newblock \href {https://doi.org/10.18653/v1/2023.emnlp-demo.46} {{O}mni{E}vent: A comprehensive, fair, and easy-to-use toolkit for event understanding}.
\newblock In \emph{Proceedings of the 2023 Conference on Empirical Methods in Natural Language Processing: System Demonstrations}, pages 508--517, Singapore. Association for Computational Linguistics.

\bibitem[{Peng et~al.(2023{\natexlab{b}})Peng, Wang, Yao, Zeng, Hou, Li, Liu, and Shen}]{peng-etal-2023-devil}
Hao Peng, Xiaozhi Wang, Feng Yao, Kaisheng Zeng, Lei Hou, Juanzi Li, Zhiyuan Liu, and Weixing Shen. 2023{\natexlab{b}}.
\newblock \href {https://doi.org/10.18653/v1/2023.findings-acl.586} {The devil is in the details: On the pitfalls of event extraction evaluation}.
\newblock In \emph{Findings of the Association for Computational Linguistics: ACL 2023}, pages 9206--9227, Toronto, Canada. Association for Computational Linguistics.

\bibitem[{Pouran Ben~Veyseh et~al.(2022)Pouran Ben~Veyseh, Ebrahimi, Dernoncourt, and Nguyen}]{pouran-ben-veyseh-etal-2022-mee}
Amir Pouran Ben~Veyseh, Javid Ebrahimi, Franck Dernoncourt, and Thien Nguyen. 2022.
\newblock \href {https://doi.org/10.18653/v1/2022.emnlp-main.652} {{MEE}: A novel multilingual event extraction dataset}.
\newblock In \emph{Proceedings of the 2022 Conference on Empirical Methods in Natural Language Processing}, pages 9603--9613, Abu Dhabi, United Arab Emirates. Association for Computational Linguistics.

\bibitem[{Satyapanich et~al.(2020)Satyapanich, Ferraro, and Finin}]{Satyapanich_Ferraro_Finin_2020}
Taneeya Satyapanich, Francis Ferraro, and Tim Finin. 2020.
\newblock \href {https://doi.org/10.1609/aaai.v34i05.6401} {Casie: Extracting cybersecurity event information from text}.
\newblock \emph{Proceedings of the AAAI Conference on Artificial Intelligence}, 34(05):8749--8757.

\bibitem[{Sun et~al.(2022)Sun, Li, Pergola, Wallace, John, Greene, Kim, and He}]{sun-etal-2022-phee}
Zhaoyue Sun, Jiazheng Li, Gabriele Pergola, Byron Wallace, Bino John, Nigel Greene, Joseph Kim, and Yulan He. 2022.
\newblock \href {https://doi.org/10.18653/v1/2022.emnlp-main.376} {{PHEE}: A dataset for pharmacovigilance event extraction from text}.
\newblock In \emph{Proceedings of the 2022 Conference on Empirical Methods in Natural Language Processing}, pages 5571--5587, Abu Dhabi, United Arab Emirates. Association for Computational Linguistics.

\bibitem[{Wang et~al.(2023)Wang, Liang, Meng, Sun, Shi, Li, Xu, Qu, and Zhou}]{wang2023chatgpt}
Jiaan Wang, Yunlong Liang, Fandong Meng, Zengkui Sun, Haoxiang Shi, Zhixu Li, Jinan Xu, Jianfeng Qu, and Jie Zhou. 2023.
\newblock Is chatgpt a good nlg evaluator? a preliminary study.
\newblock \emph{arXiv preprint arXiv:2303.04048}.

\bibitem[{Wang et~al.(2022)Wang, Yu, Chang, Sun, and Huang}]{wang-etal-2022-query}
Sijia Wang, Mo~Yu, Shiyu Chang, Lichao Sun, and Lifu Huang. 2022.
\newblock \href {https://doi.org/10.18653/v1/2022.findings-acl.16} {Query and extract: Refining event extraction as type-oriented binary decoding}.
\newblock In \emph{Findings of the Association for Computational Linguistics: ACL 2022}, pages 169--182, Dublin, Ireland. Association for Computational Linguistics.

\bibitem[{Wei et~al.(2024)Wei, Cui, Cheng, Wang, Zhang, Huang, Xie, Xu, Chen, Zhang, Jiang, and Han}]{wei2024chatiezeroshotinformationextraction}
Xiang Wei, Xingyu Cui, Ning Cheng, Xiaobin Wang, Xin Zhang, Shen Huang, Pengjun Xie, Jinan Xu, Yufeng Chen, Meishan Zhang, Yong Jiang, and Wenjuan Han. 2024.
\newblock \href {https://arxiv.org/abs/2302.10205} {Chatie: Zero-shot information extraction via chatting with chatgpt}.
\newblock \emph{Preprint}, arXiv:2302.10205.

\bibitem[{Wei and Wang(2019)}]{surveyofee}
Xiang Wei and Bang Wang. 2019.
\newblock \href {https://doi.org/10.1109/ACCESS.2019.2956831} {A survey of event extraction from text}.
\newblock \emph{IEEE Access}, PP:1--1.

\bibitem[{Yan et~al.(2023)Yan, Sun, Li, Zhou, Huang, and Qiu}]{yan-etal-2023-utc}
Hang Yan, Yu~Sun, Xiaonan Li, Yunhua Zhou, Xuanjing Huang, and Xipeng Qiu. 2023.
\newblock \href {https://doi.org/10.18653/v1/2023.acl-long.226} {{UTC}-{IE}: A unified token-pair classification architecture for information extraction}.
\newblock In \emph{Proceedings of the 61st Annual Meeting of the Association for Computational Linguistics (Volume 1: Long Papers)}, pages 4096--4122, Toronto, Canada. Association for Computational Linguistics.

\bibitem[{Yang et~al.(2024)Yang, Yang, Hui, Zheng, Yu, Zhou, Li, Li, Liu, Huang, Dong, Wei, Lin, Tang, Wang, Yang, Tu, Zhang, Ma, Yang, Xu, Zhou, Bai, He, Lin, Dang, Lu, Chen, Yang, Li, Xue, Ni, Zhang, Wang, Peng, Men, Gao, Lin, Wang, Bai, Tan, Zhu, Li, Liu, Ge, Deng, Zhou, Ren, Zhang, Wei, Ren, Liu, Fan, Yao, Zhang, Wan, Chu, Liu, Cui, Zhang, Guo, and Fan}]{yang2024qwen2technicalreport}
An~Yang, Baosong Yang, Binyuan Hui, Bo~Zheng, Bowen Yu, Chang Zhou, Chengpeng Li, Chengyuan Li, Dayiheng Liu, Fei Huang, Guanting Dong, Haoran Wei, Huan Lin, Jialong Tang, Jialin Wang, Jian Yang, Jianhong Tu, Jianwei Zhang, Jianxin Ma, Jianxin Yang, Jin Xu, Jingren Zhou, Jinze Bai, Jinzheng He, Junyang Lin, Kai Dang, Keming Lu, Keqin Chen, Kexin Yang, Mei Li, Mingfeng Xue, Na~Ni, Pei Zhang, Peng Wang, Ru~Peng, Rui Men, Ruize Gao, Runji Lin, Shijie Wang, Shuai Bai, Sinan Tan, Tianhang Zhu, Tianhao Li, Tianyu Liu, Wenbin Ge, Xiaodong Deng, Xiaohuan Zhou, Xingzhang Ren, Xinyu Zhang, Xipin Wei, Xuancheng Ren, Xuejing Liu, Yang Fan, Yang Yao, Yichang Zhang, Yu~Wan, Yunfei Chu, Yuqiong Liu, Zeyu Cui, Zhenru Zhang, Zhifang Guo, and Zhihao Fan. 2024.
\newblock \href {https://arxiv.org/abs/2407.10671} {Qwen2 technical report}.
\newblock \emph{Preprint}, arXiv:2407.10671.

\bibitem[{Zhang et~al.(2020)Zhang, Liu, Pan, Song, and Leung}]{10.1145/3366423.3380107}
Hongming Zhang, Xin Liu, Haojie Pan, Yangqiu Song, and Cane Wing-Ki Leung. 2020.
\newblock \href {https://doi.org/10.1145/3366423.3380107} {Aser: A large-scale eventuality knowledge graph}.
\newblock In \emph{Proceedings of The Web Conference 2020}, WWW '20, page 201–211, New York, NY, USA. Association for Computing Machinery.

\bibitem[{Zheng et~al.(2023)Zheng, Chiang, Sheng, Zhuang, Wu, Zhuang, Lin, Li, Li, Xing, Zhang, Gonzalez, and Stoica}]{zheng2023judgingllmasajudgemtbenchchatbot}
Lianmin Zheng, Wei-Lin Chiang, Ying Sheng, Siyuan Zhuang, Zhanghao Wu, Yonghao Zhuang, Zi~Lin, Zhuohan Li, Dacheng Li, Eric~P. Xing, Hao Zhang, Joseph~E. Gonzalez, and Ion Stoica. 2023.
\newblock \href {https://arxiv.org/abs/2306.05685} {Judging llm-as-a-judge with mt-bench and chatbot arena}.
\newblock \emph{Preprint}, arXiv:2306.05685.

\bibitem[{Zheng et~al.(2021)Zheng, Cao, Xu, and Bian}]{zheng-etal-2021-revisiting}
Shun Zheng, Wei Cao, Wei Xu, and Jiang Bian. 2021.
\newblock \href {https://doi.org/10.18653/v1/2021.findings-acl.405} {Revisiting the evaluation of end-to-end event extraction}.
\newblock In \emph{Findings of the Association for Computational Linguistics: ACL-IJCNLP 2021}, pages 4609--4617, Online. Association for Computational Linguistics.

\end{thebibliography}

\appendix

\section{Datasets Statistics}
\label{sec:datasets}
We utilize 10 widely-used datasets as evaluation benchmarks in both EM and RAEE evaluation. Since these datasets originate from various domains and differ in terms of pre-defined ontologies as well as size, it is necessary to introduce basic statistics of these datasets, as shown in Table~\ref{tab:datasets}.

\begin{table}[h]
    \centering
    \small
    \begin{threeparttable}
        \begin{tabular}{l|ccccc}
            \toprule
               Dataset   & \#Inst & \#ET & \#Evt & \#RT & \#Arg \\
            \midrule
            \multicolumn{6}{l}{\textit{ED and EAE tasks}} \\
            \midrule
               ACE05-en   & 20920 & 33 & 5348 & 22 & 8097 \\
               Genia2011  & 1375 & 9 & 13537 & 10 & 11865 \\
               CASIE  & 1483 & 5 & 8469 & 26 & 22575 \\
               PHEE  & 4827 & 2 & 5019 & 16 & 25760 \\
            \midrule
            \multicolumn{6}{l}{\textit{ED task only}} \\
            \midrule
               FewEvent  & 12573 & 100 & 12573 & — & — \\
               MEE-en & 13000 & 16 & 17257 & — & — \\
               M$^2$E$^2$  & 6013 & 8 & 1105 & — & — \\
            \midrule
            \multicolumn{6}{l}{\textit{EAE task only}} \\
            \midrule
               WikiEvents  & 565 & 50 & 3932 & 58 & 5501 \\
               RAMS  & 9647 & 139 & 9647 & 65 & 21206 \\
               GENEVA  & 3684 & 115 & 7505 & 220 & 12314 \\
            \bottomrule
        \end{tabular}
    \end{threeparttable}
    \caption{Statistics of 10 datasets used in RAEE. \#Inst, \#ET, \#Evt, \#RT, and \#Arg represent the number of data instances, event type, event instances, role types, and argument instances, respectively.}
    \label{tab:datasets}
\end{table}

\section{Key Criteria in RAEE}
\label{sec:key_criteria}

Considering that the requirements for EE vary across different scenarios and applications, we introduce key criteria in RAEE to enable the adaptive mechanism. As shown in Tabel~\ref{tab:key_criteria}, these key criteria are based on factors that significantly influence the evaluation results in our human annotation meta-evaluation experiment. Furthermore, this adaptive mechanism is easily extendable, allowing for the addition of specific judgment criteria to account for more factors in extraction.

\begin{table*}[h]
    \centering
    \small
    \begin{threeparttable}
        \begin{tabularx}{\textwidth}{X|c}
        \toprule
        \textbf{Key Criteria Template} & \textbf{Setting} \\ 
        \midrule
           If there is a more accurate predefined type for a predicted trigger/argument, then only the more accurate type can be reassessed as \{\} one  & correct \\
        \midrule
           If a predict trigger/argument is more reasonable than golden annotations, it should be reassessed as \{\} one & correct \\
        \midrule
           When there is co-reference, pronouns can be reassessed as \{\} one & correct \\
        \midrule
           If the core word is correct but some modifiers are missing, the predicted trigger/argument should be reassessed as \{\} one & correct \\
        \midrule
           A predicted trigger is reassessed as \{\} one when it triggers an event that does not really occur & incorrect \\
        \midrule
           A golden trigger/argument is reassessed as \{\} one if there is no correspondence in predicted triggers/arguments, even if there is a predicted trigger/argument with a more accurate or reasonable type & not recalled \\
        \midrule
           When there is co-reference, pronouns can be reassessed as \{\} one & recalled \\
        \midrule
           If the core word is recalled but some modifiers are missing, the golden trigger/argument should be reassessed as \{\} one & recalled \\
        \bottomrule
        \end{tabularx}
    \end{threeparttable}
    \caption{Key criteria templates and their settings used in RAEE.}
    \label{tab:key_criteria}
\end{table*}

\section{Prompts}
\subsection{Reassessment Prompts in RAEE}
\label{sec:reassess_prompts}

Table~\ref{tab:reassess_prompts_ed_precision}, \ref{tab:reassess_prompts_ed_recall}, \ref{tab:reassess_prompts_eae_precision} and \ref{tab:reassess_prompts_eae_recall} illustrate the prompts we use for reassessment in RAEE.

\subsection{Inference Prompts for LLMs Extracting Events}
\label{sec:inference_prompts}

Table~\ref{tab:inference_prompts_ed} and \ref{tab:inference_prompts_eae} illustrate the prompts we use for testing the event extraction ability of LLMs in ED and EAE tasks, respectively.
It should be noted that we follow existing closed domain EE studies \cite{gao2023exploringfeasibilitychatgptevent, wei2024chatiezeroshotinformationextraction, cai-etal-2024-improving-event} to explicitly provide all possible types in the inference prompt. This is a necessary step when performing EE inference with LLMs and does not constitute information leakage.

\begin{table*}[h]
    \centering
    \small
    \begin{threeparttable}
    \begin{tabularx}{\textwidth}{cX}
     \toprule
        \multicolumn{2}{l}{\textbf{Prompt Used for Reassessing on ED Task and Precision Metric}} \\
     \midrule
        \multirow{5}{*}{Instruction} & You are a reassessor designed to reassess predicted event triggers from an event detection model. Task Description: Refer to a piece of context and golden triggers, reassess predicted event triggers. A trigger is the key word in the sentence that most explicitly conveys the occurrence of the event. A predicted trigger can be reassessed as correct one if it conveys the occurrence of a particular interest kind of event, even if it does not exactly match any of golden triggers; otherwise, it is reassessed as incorrect one. \\
        ~ \\
        \multirow{6}{*}{Key Criteria} & There are several rules to follow when reassessing: (1) If there is a more accurate predefined type for a predicted trigger, then only the more accurate type can be reassessed as correct one; (2) If a predict trigger is more reasonable than golden annotations, it should be reassessed as correct one; (3) When there is co-reference, pronouns can be reassessed as correct one; (4) If the core word is correct but some modifiers are missing, the predicted trigger should be reassessed as correct one; (5) A predicted trigger is reassessed as incorrect one when it triggers an event that does not really occur. \\
        ~ \\
        \multirow{6}{*}{Requirements} & When answering, provide your thought process at first, including analyzing the context, explaining how golden triggers indicate events with a specific type, and reassessing predicted triggers one by one based on your thought. After that, answer final reassessment results. \\
        ~ & Please answer in the following python dict format: \{'Thought Process': \{'Context Analysis': str, 'Gold Triggers': [your explanation for gold trigger 0, your explanation for gold trigger 1, ...], 'Predicted Triggers': [your reassessment thought for predicted trigger 0, your reassessment thought for predicted trigger 1, ...]\}, 'Final Reassessment Results': [1, 0, ...]\}.  \\
        ~ \\
        \multirow{14}{*}{Data} & Events of interest include: \\
        ~ & \quad Contact.Meet. The event is related to contact meet. \\
        ~ & \quad Business.Start-Org. The event is related to business start organization. \\
        ~ & Context: \\
        ~ & \quad Professor Massimo Pigliucci , a [t.Pred.1] [t.Pred.0] professor [/t.Pred.0] [/t.Pred.1] of ecology and evolution at the State University of New York at Stony Brook , has criticized ICR for professing to present the same science as that taught in secular universities while at the same time requiring students and faculty to sign a statement of faith to ICR 's fundamentalist religious mission , most notably in affirming conformity in all its [t.Gold.0] work [/t.Gold.0] to Biblical doctrine . \\
        ~ & The position of each following trigger is labeled in the context between a pair of [t] and [/t]. \\
        ~ & Trigger Gold.0: work \# event type: Business.Employment-Tenure \\
        ~ & Trigger Pred.0: professor \# event type: Business.Employment-Tenure\\
        ~ & Trigger Pred.1: professor \# event type: Education.Education \\
     \bottomrule
    \end{tabularx}
    \end{threeparttable}
    \caption{Prompt used for reassessing on ED task and precision metric. An example of data is illustrated.}
    \label{tab:reassess_prompts_ed_precision}
\end{table*}

\begin{table*}[h]
    \centering
    \small
    \begin{threeparttable}
    \begin{tabularx}{\textwidth}{cX}
     \toprule
        \multicolumn{2}{l}{\textbf{Prompt Used for Reassessing on ED Task and Recall Metric}} \\
     \midrule
        \multirow{6}{*}{Instruction} & You are a reassessor designed to reassess the recall result of an event detection model. Task Description: According to a piece of context, predicted triggers and golden triggers, reassess whether each golden trigger is recalled. A trigger is the key word in the sentence that most explicitly conveys the occurrence of the event. A golden trigger can be reassessed as recalled one if it can correspond to one or more predicted triggers that convey the occurrence of an event with the same interest type, even if it does not exactly match any of predicted triggers; otherwise, it is reassessed as not recalled one. \\
        ~ \\
        \multirow{4}{*}{Key Criteria} & There are several rules to follow when reassessing: (1) A golden trigger is reassessed as not recalled one if there is no correspondence in predicted triggers, even if there is a predicted trigger with more accurate or reasonable type; (2) When there is co-reference, pronouns can be reassessed as recalled one; (3) If the core word is recalled but some modifiers are missing, the golden trigger should be reassessed as recalled one. \\
        ~ \\
        \multirow{7}{*}{Requirements} & When answering, provide your thought process at first, including analyzing the context, explaining how golden triggers indicate events with a specific type, finding correspondence with predicted triggers for each golden trigger or explaining why any correspondence cannot be found, and reassessing golden triggers one by one based on your thought. After that, answer final reassessment results. \\
        ~ & Please answer in the following python dict format: \{'Thought Process': \{'Context Analysis': str, 'Gold Triggers': [your finding or explanation for gold trigger 0, your finding or explanation for gold trigger 1, ...]\}, 'Final Reassessment Results': [1 , 0 , ...]\}.  \\
        ~ \\
        \multirow{9}{*}{Data} & Events of interest include: \\
        ~ & \quad Business.Start-Subsidiary. The event is related to business start subsidiary. \\
        ~ & Context: \\
        ~ & \quad A subsidiary company or daughter company is a company that is [t.Gold.0] owned or [t.Pred.0] controlled [/t.Pred.0] by [/t.Gold.0] another company , which is called the parent company , parent , or holding company . \\
        ~ & The position of each following trigger is labeled in the context between a pair of [t] and [/t]. \\
        ~ & Trigger Pred.0: controlled \# event type: Business.Start-Subsidiary \\
        ~ & Trigger Gold.0: owned or controlled by \# event type: Business.Start-Subsidiary \\
     \bottomrule
    \end{tabularx}
    \end{threeparttable}
    \caption{Prompt used for reassessing on ED task and recall metric. An example of data is illustrated.}
    \label{tab:reassess_prompts_ed_recall}
\end{table*}

\begin{table*}[h]
    \centering
    \small
    \begin{threeparttable}
    \begin{tabularx}{\textwidth}{cX}
    \toprule
        \multicolumn{2}{l}{\textbf{Prompt Used for Event Detection}} \\
    \midrule
         \multirow{4}{*}{Instruction} & You are an event extractor designed to check for the presence of a specific event in a piece of context and to locate the corresponding event trigger. Task Description: Identify all triggers related to the event of interest in the context. A trigger is the key word in the context that most explicitly conveys the occurrence of the event. \\
         ~ & \\
         \multirow{5}{*}{Requirements} & When extracting, analyzing the context at first. After that, answer extraction results. You need to extract the span of a extracted trigger as well as its corresponding event type. Note that there may be zero to plural triggers in the context. Please answer in the following python dict format: \{'Context Analysis': str, 'Extraction Results': [\{'Trigger Span': a span in the context, 'Event Type': a specific event of interest\}, ...]\}. \\
         ~ & \\
         \multirow{7}{*}{Data} & Types of events that may occur include: \\
         ~ & \quad Adverse\_event. The event is related to adverse event. \\
         ~ & \quad Potential\_therapeutic\_event. The event is related to potential therapeutic event. \\
         ~ & Context: \\
         ~ & \quad The case histories are presented of two patients who developed lung disease associated with the use of nitrofurantoin with histological features of bronchiolitis obliterans organising pneumonia ( BOOP ) , a rare but recognised form of drug induced injury . \\
    \bottomrule
    \end{tabularx}
    \end{threeparttable}
    \caption{Prompt used for testing the event extraction ability of LLMs in ED task. An example of data is illustrated.}
    \label{tab:inference_prompts_ed}
\end{table*}

\begin{table*}[h]
    \centering
    \small
    \begin{threeparttable}
    \begin{tabularx}{\textwidth}{cX}
     \toprule
        \multicolumn{2}{l}{\textbf{Prompt Used for Reassessing on EAE Task and Precision Metric}} \\
     \midrule
        \multirow{6}{*}{Instruction} & You are a reassessor designed to reassess predicted event arguments from an event argument extraction model. Task Description: Refer to a piece of context and golden arguments, reassess predicted event arguments. Arguments have the semantic role corresponding to the given event trigger by the word span between [t] and [/t]. A predicted argument can be reassessed as correct one if it have a particular interest type of role corresponding to the given event trigger, even if it does not exactly match any of golden arguments; otherwise, it is reassessed as incorrect one. \\
        ~ \\
        \multirow{6}{*}{Key Criteria} & There are several rules to follow when reassessing: (1) If there is a more accurate predefined role type for a predicted argument, then only the more accurate role type can be reassessed as correct one; (2) If a predict argument is more reasonable than golden annotations, it should be reassessed as correct one; (3) When there is co-reference, pronouns can be reassessed as correct one; (4) If the core word is correct but some modifiers are missing, the predicted argument should be reassessed as correct one. \\
        ~ \\
        \multirow{7}{*}{Requirements} & When answering, provide your thought process at first, including analyzing the context, explaining roles of golden arguments in the event of interest, and reassessing predicted arguments one by one based on your thought. After that, answer final reassessment results. \\
        ~ & Please answer in the following python dict format: \{'Thought Process': \{'Context Analysis': str, 'Gold Arguments': [your explanation for gold argument 0, your explanation for gold argument 1, ...], 'Predicted Arguments': [your reassessment thought for predicted argument 0, your reassessment thought for predicted argument 1, ...]\}, 'Final Reassessment Results': [1, 0, ...]\}.   \\
        ~ \\
        \multirow{6}{*}{Data} & The event of interest is Justice:Trial-Hearing. The event is related to a trial or hearing for someone. \\
        ~ & Argument roles of interest: ['Defendant', 'Prosecutor', 'Place', 'Adjudicator'] \\
        ~ & Context: \\
        ~ & \quad She was not [t] tried [/t] in the deaths of the other two . \\
        ~ & Gold argument 0: She \# role type: Defendant \\
        ~ & Pred argument 0: deaths \# role type: Defendant \\
     \bottomrule
    \end{tabularx}
    \end{threeparttable}
    \caption{Prompt used for reassessing on EAE task and precision metric. An example of data is illustrated.}
    \label{tab:reassess_prompts_eae_precision}
\end{table*}

\begin{table*}[h]
    \centering
    \small
    \begin{threeparttable}
    \begin{tabularx}{\textwidth}{cX}
    \toprule
        \multicolumn{2}{l}{\textbf{Prompt Used for Event Argument Extraction}} \\
    \midrule
         \multirow{3}{*}{Instruction} & You are an argument extractor designed to check for the presence of arguments regarding specific roles for an event in a piece of context. Task Description: Identify all event arguments related to roles of interest in the context. These arguments should have the semantic role corresponding to the given event trigger by the word span between [t] and [/t].  \\
         ~ & \\
         \multirow{4}{*}{Requirements} & When extracting, analyzing the context at first. After that, answer extraction results. Note that there may be zero to plural arguments for each role. Please answer in the following python dict format: \{'Context Analysis': str, 'Extraction Results': \{'role1': [str(the argument span), ...(The length of this list depends on the number of arguments that you extract for this kind of role)], 'role2': [str, ...], ...\}\}. \\
         ~ & \\
         \multirow{11}{*}{Data} & The event of interest is justice.initiatejudicialprocess.trialhearing. The event is related to justice initiative judicial process trial hearing. \\
         ~ & Roles of interest include: prosecutor, defendant, judgecourt, crime, place. \\
         ~ & Context: \\
         ~ & \quad But no longer . By the end of 2015 the establishment was expressing considerable dismay and desperation over its inability to do so , as the Republican base and its choices fell out of control . Republican elected officials and contenders for the next presidential election expressed open contempt for the Paris deliberations , refusing to even attend the proceedings . The three candidates who led in the polls at the time Donald Trump , Ted Cruz , and Ben Carson  adopted the stand of the largely evangelical base : humans have no impact on global warming , if it is happening at all . The other candidates reject government action to deal with the matter . \\
    \bottomrule
    \end{tabularx}
    \end{threeparttable}
    \caption{Prompt used for testing the event extraction ability of LLMs in EAE task. An example of data is illustrated.}
    \label{tab:inference_prompts_eae}
\end{table*}

\section{Implementation Details}
\label{sec:implementation_details}
To avoid inconsistency in evaluation caused by factors out of model frameworks such as data preprocessing or Python package versions, we follow TextEE \cite{huang2024texteebenchmarkreevaluationreflections}, utilize its standardized data splits and unified input-output framework for evaluating models, and adhere to its default training configs, which can be found in the corresponding GitHub repository. All discriminative models are based on the RoBERTa-large \cite{liu2019robertarobustlyoptimizedbert} pre-trained model, and generative models are built using BART-large \cite{lewis2019bartdenoisingsequencetosequencepretraining}.

\begin{table*}[t]
    \begin{threeparttable}
    \small
    \begin{tabularx}{\textwidth}{cX}
     \toprule
        \multicolumn{2}{l}{\textbf{Prompt Used for Reassessing on EAE Task and Recall Metric}} \\
     \midrule
        \multirow{7}{*}{Instruction} & You are a reassessor designed to reassess the recall result of an event argument extraction model. Task Description: According to a piece of context, predicted arguments and golden arguments, reassess whether each golden argument is recalled. Arguments have the semantic role corresponding to the given event trigger by the word span between [t] and [/t]. A golden argument can be reassessed as recalled one if it can correspond to one or more predicted arguments that have the same interest type of role corresponding to the given event trigger, even if it does not exactly match any of predicted arguments; otherwise, it is reassessed as not recalled one. \\
        ~ \\
        \multirow{5}{*}{Key Criteria} & There are several rules to follow when reassessing: (1) A golden argument is reassessed as not recalled one if there is no correspondence in predicted arguments, even if there is a predicted argument with more accurate or reasonable role type; (2) When there is co-reference, pronouns can be reassessed as recalled one; (3) If the core word is recalled but some modifiers are missing, the golden argument should be reassessed as recalled one. \\
        ~ \\
        \multirow{7}{*}{Requirements} & When reassessing, provide your thought process at first, including analyzing the context, explaining roles of golden arguments in the event of interest, finding correspondence with predicted arguments for each golden argument or explaining why any correspondence cannot be found, and reassessing golden arguments one by one based on your thought. After that, answer final reassessment results. \\
        ~ & Please answer in the following python dict format: \{'Thought Process': \{'Context Analysis': str, 'Gold Arguments': [your finding or explanation for gold argument 0, your finding or explanation for gold argument 1, ...]\}, 'Final Reassessment Results': [1, 0, ...]\}. \\
        ~ \\
        \multirow{7}{*}{Data} & The event of interest is Life:Die. The event is related to life and someone died. \\
        ~ & Argument roles of interest: ['Agent', 'Victim', 'Instrument', 'Place'] \\
        ~ & Context: \\
        ~ & \quad There 's nothing very pretty about the training that you take to prepare you for combat , because it is to [t] kill [/t] people . \\
        ~ & Pred argument 0: people \# role type: Victim \\
        ~ & Gold argument 0: you \# role type: Agent \\
     \bottomrule
    \end{tabularx}
    \end{threeparttable}
    \caption{Prompt used for reassessing on EAE task and recall metric. An example of data is illustrated.}
    \label{tab:reassess_prompts_eae_recall}
\end{table*}
\end{document}